\documentclass[runningheads]{llncs}

\usepackage{eccv}

\usepackage{eccvabbrv}

\usepackage{graphicx}
\usepackage{booktabs}

\usepackage[accsupp]{axessibility}  

\usepackage[pagebackref,breaklinks,colorlinks,citecolor=eccvblue]{hyperref}

\usepackage{orcidlink}
\usepackage{tabularx}
\usepackage{multirow}
\usepackage{tcolorbox}
\usepackage{makecell,microtype}
\usepackage{tcolorbox}
\usepackage{wrapfig}
\usepackage[normalem]{ulem}
\usepackage[table, dvipsnames]{xcolor}

\usepackage{soul}
\usepackage{pifont}
\usepackage{subcaption}
\usepackage{float}
\tcbuselibrary{listings, skins}
\usepackage{listings}
\usepackage{caption}
\usepackage{array}

\DeclareCaptionLabelFormat{andfigure}{#1~#2~\&~\figurename~\thefigure}
\DeclareCaptionLabelFormat{andtable}{#1~#2~\&~\tablename~\thetable}

\definecolor{promptGray}{RGB}{247,247,250}
\definecolor{speakerBlue}{RGB}{22,82,172}
\definecolor{listenerPurple}{RGB}{108,28,155}
\definecolor{inferGreen}{RGB}{14,122,92}
\definecolor{headerRule}{RGB}{210,210,220}

\lstdefinestyle{promptstyle}{
  basicstyle=\small\ttfamily,
  columns=fullflexible,
  keepspaces=true,
  breaklines=true,
  breakatwhitespace=true,
  breakindent=12pt,
  showstringspaces=false,
  frame=none,
  aboveskip=0pt,
  belowskip=0pt,
  xleftmargin=0pt,
  resetmargins=true,
}

\tcbset{
  promptbase/.style={
    enhanced,
    colback=promptGray,
    left=10pt, right=10pt, top=6pt, bottom=8pt,
    arc=5pt,
    boxrule=0.8pt,
    attach boxed title to top left={xshift=12pt, yshift=-\tcboxedtitleheight/2},
    boxed title style={
      sharp corners, left=5pt, right=5pt, top=2pt, bottom=2pt, boxrule=0pt,
    },
    fonttitle=\small\bfseries\sffamily,
    coltitle=white,
  },
  speakerstyle/.style={
    promptbase, colframe=speakerBlue,
    boxed title style={colback=speakerBlue, sharp corners},
  },
  listenerstyle/.style={
    promptbase, colframe=listenerPurple,
    boxed title style={colback=listenerPurple, sharp corners},
  },
  inferstyle/.style={
    promptbase, colframe=inferGreen,
    boxed title style={colback=inferGreen, sharp corners},
  },
}

\definecolor{ttt}{HTML}{E8F3FF}        
\definecolor{pretrain}{HTML}{FFF0E6}   
\definecolor{tfree}{HTML}{EEFAEE} 

\definecolor{wrongred}{RGB}{248,206,204}
\definecolor{ours}{HTML}{F3E8FF} 

\sethlcolor{wrongred}

\tcbset{
  colback=gray!4,
  colframe=gray!30,
  boxrule=0.4pt,
  arc=2pt,
  left=6pt,
  right=6pt,
  top=6pt,
  bottom=6pt,
}

\newcommand{\vlm}{MLLM\xspace}
\newcommand{\vlms}{MLLMs\xspace}
\newcommand{\ourmethod}{RRG\xspace}
\newcommand{\ourmethodlong}{\textbf{R}einforced \textbf{R}eference \textbf{G}ame\xspace}
\newcommand{\yollava}{Yo’LLaVA\xspace}

\renewcommand{\paragraph}[1]{\smallskip \noindent\textbf{#1}}
\newcommand{\suppmat}{\textit{Supp. Mat.}}

\newcommand{\inlineColorbox}[2]{\begingroup\setlength{\fboxsep}{1pt}\colorbox{#1}{\hspace*{2pt}\vphantom{Ay}#2\hspace*{2pt}}\endgroup}

\newcolumntype{Y}{>{\centering\arraybackslash}X} 
\newcolumntype{C}[1]{>{\centering\arraybackslash}m{#1}} 
\newcolumntype{L}[1]{>{\raggedright\arraybackslash}m{#1}} 
\newcolumntype{R}[1]{>{\raggedleft\arraybackslash}m{#1}} %

\newcolumntype{C}[1]{>{\centering\arraybackslash}m{#1}}

\begin{document}

\title{Personalizing MLLMs via Reinforced Multimodal Reference Game} 


\author{Deepayan Das\inst{1}\orcidlink{0009-0007-5285-6520} \and
Davide Talon\inst{2}\orcidlink{0009-0003-6029-1532} \and
Yiming Wang\inst{2}\orcidlink{0000-0002-5932-4371} \and
Massimiliano Mancini\inst{1}\orcidlink{0000-0001-8595-9955} \and
Elisa Ricci\inst{1, 2}\orcidlink{0000-0002-0228-1147}}

\authorrunning{D.~Das et al.}

\institute{University of Trento, Trento, Italy \\
\email{\{deepayan.das, massimiliano.mancini, e.ricci\}@unitn.it} \and
Fondazione Bruno Kessler, Trento, Italy\\
\email{\{dtalon,ywang\}@fbk.eu}} 

\maketitle

\begin{abstract}
Personalizing Multimodal Large Language Models (MLLMs) aims to recognize users’ unique concepts from visual data and provide personalized responses. Although prior work has shown the benefit of concept descriptions and reasoning for this task, MLLM descriptions often include information, such as state and context, that does not help and may in fact hinder the unique identification of the target concept among other visually similar items. Effective descriptions of personal concepts should instead be accurate, discriminative, and free of distracting details. To achieve such descriptions, we introduce Reinforced Reference Game (RRG), a learning framework that promotes discriminative descriptions through a novel reinforced multimodal reference game. The MLLM plays both the roles of speaker and listener in a contrastive game setting, whose goal is to effectively communicate discriminative information about a target concept. Our approach formulates a verifiable contrastive reward over hard positives (dissimilar views of the same concept) and hard negatives (visually similar but different concepts). Empirically, RRG achieves state-of-the-art across multiple tasks on three personalization benchmarks.
RRG generalizes to unseen domains and outperforms existing methods based on concept descriptions and personalization-specific RL frameworks. We will release code and models in the \href{https://deepayan137.github.io/papers/conversational-personalization.html}{project page}.
\end{abstract}    
\section{Introduction}
\label{sec:intro}

Personalization~\cite{alaluf2024myvlm, nguyen2024yollava, hao2025remember, das2025training} aims to equip Multimodal Large Language Models (MLLMs)~\cite{liu2023visual, achiam2023gpt, wang2024qwen2, alayrac2022flamingo, li2023blip} with knowledge about user-defined concepts~\cite{alaluf2024myvlm, nguyen2024yollava}. In this way, an MLLM can answer questions, describe, or even refer to such concepts, providing outputs that are more engaging and aligned with user's expectations.

\begin{figure}[t!]

\centering
\includegraphics[width=\linewidth]{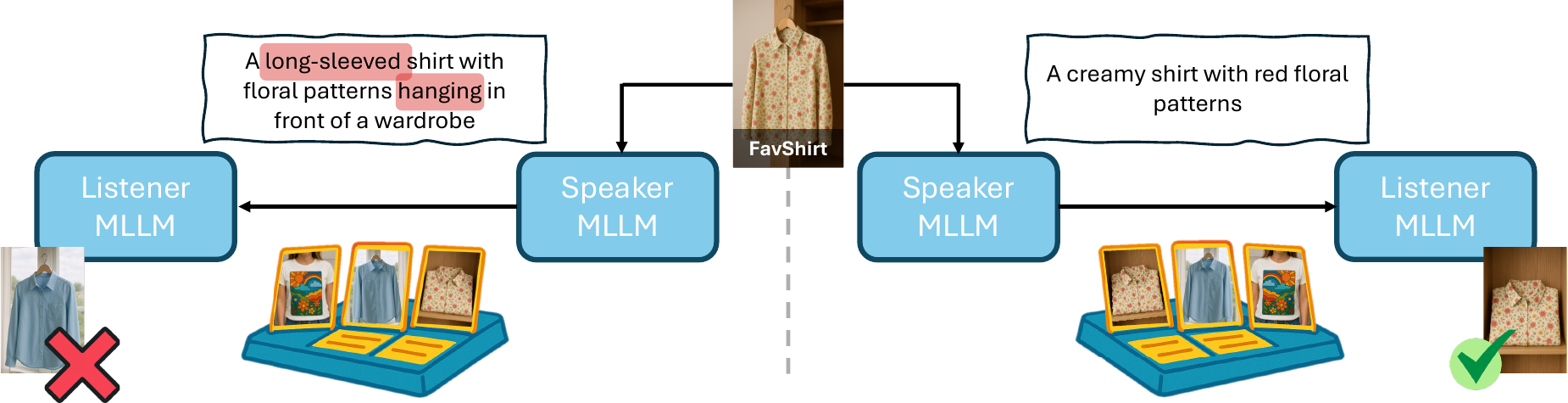}
    \caption{\textbf{The idea behind our \ourmethod.} We frame personalization as a reference game. The MLLM plays both the role of the \textit{speaker} (describing a target concept) and the \textit{listener} (identifying it among candidates), which encourages the speaker to generate distinctive, invariant descriptions that enable accurate identification beyond standard dense captioning. On the left a failing round, on the right a successful one.}
    \label{fig:teaser}
\end{figure}

MLLM personalization entails several challenges. Consider the example in \cref{fig:teaser}: how can we distinguish Amy's {shirt} from other shirts? One could argue that Amy's shirt can be recognized from its {red floral patterns and creamy color}. Note that these attributes (i) provide a unique, fine-grained identification of the item {and (ii) the description does not have distracting information, \ie, attributes are limited to the user-concept itself and do not change over time.}

To equip MLLMs with this capability, early personalization approaches learn concept-specific representations at test-time~\cite{alaluf2024myvlm, nguyen2024yollava, nguyen2025yo}. However, these methods are computationally expensive and require retraining whenever new concepts are introduced. Recent works sidestep these limitations by either reasoning over reference images~\cite{hao2025remember, seifi2025personalization, oh2025repic} or by introducing template-based textual descriptions~\cite{das2025training}. Given a query image, such models retrieve the closest personal concepts in a database and reason over them directly during inference.
While effective, these strategies either rely on the model’s ability to identify which regions of the reference images are relevant or assume descriptions can accurately focus on the concept of interest. As a result, concept representations that contain irrelevant or mutable details can mislead the recognition of personal concepts.

Why do standard descriptions fall short? Consider Amy's personal long-sleeved t-shirt with floral patterns in~\cref{fig:teaser}. First, descriptions frequently mention the \textit{concept state}: visual features may expose the current state of the concept, \eg, ``hanging'', which is irrelevant for recognition and may confuse the model either when the state of the shirt changes or when other similar-looking shirts share the same state. Secondly, they include \textit{context information} such as background or positional cues which can vary with viewpoint changes and thereby can hinder generalization. Finally, regular descriptions lack \textit{comparative focus}, \ie, they ignore what truly distinguishes the concept from semantically similar ones, for instance, the ``floral pattern'' that differentiates Amy's t-shirt from Joy's similar navy-blue one. 
As MLLMs are trained for generic captioning and VQA, they are ill-suited for the necessary fine-grained details needed for effective personalization. 

Thus, a natural question arises: \textbf{\textit{Can we teach an MLLM to communicate fine-grained yet essential concept features for personalization?}} To answer this question, we depart from previous efforts and treat personalization as a \textit{reference game}, inspired by the classic ``Guess Who?'' game (\cref{fig:teaser}). 
We introduce a novel MLLM learning paradigm, \ourmethodlong (\ourmethod), to enhance personalized concept descriptions. Specifically, given an MLLM we want to personalize, we formulate a ``game'' in which the model alternates between two roles: 
 a \textit{speaker}, which describes the target concept, and a \textit{listener}, which must identify that concept from a set of candidate images.
To succeed, the speaker must go beyond standard image captions and articulate only the distinctive and immutable features that allow the listener to select the correct concept. 
Crucially, we design the game to explicitly favor informative discriminative descriptions while discouraging distracting details, by exposing the listener to ambiguous candidate concepts that are difficult to distinguish. On the one hand, we present the speaker and the listener with two challenging views of the same concept: hard positives, which differ in background, viewpoint, or object state. On the other hand, we include distractor images composed of hard negatives, consisting of visually similar items from the same semantic class as the referenced concept.
This hard contrastive setup provides a verifiable reward signal, enabling us to train the speaker to generate discriminative and immutable descriptions. We optimize the speaker via GRPO~\cite{guo2025deepseek}, which allows learning from verifiable game outcomes, thereby eliminating the need for explicit description supervision, which is non-trivial to obtain.

Results on established benchmarks \yollava~\cite{nguyen2024yollava}, MyVLM~\cite{alaluf2024myvlm} and PerVA~\cite{das2025training} demonstrate that our trained description model effectively improves personalization performance on captioning and VQA tasks, outperforming existing training-based methods.
Notably, \ourmethod outperforms alternative description-based approaches~\cite{das2025training}, and reinforcement learning (RL)-based ones~\cite{oh2025repic}, demonstrating both the benefits of the high-quality concept descriptions and the multimodal reference game as an RL training strategy.

\noindent Our main \textbf{contributions} are summarized below:
\begin{itemize}
\item We introduce a novel perspective on \vlm  personalization, by formulating the task as a communication game, where success depends on communicating the essential, immutable attributes that uniquely identify a user-defined concept among visually similar candidates; 
\item Building on this formulation, we propose \ourmethod, a new approach in which the \vlm alternates between speaker and listener roles in a multimodal reference game, learning to generate uniquely discriminative concept descriptions. The speaker is updated through reinforcement learning, where the reward is derived from game outcomes. 
\item \ourmethod achieves state-of-the-art performance for the captioning task on three datasets and superior VQA accuracy, outperforming existing training-based methods and generalizing effectively to unseen concepts.
\end{itemize}
\section{Related Work}
\label{sec:related}
\paragraph{Personalization with \vlms.}
The problem of personalization has been explored across several vision tasks. In text-to-image generation, the idea is to bind target concepts to custom identifiers derived from a handful of reference images, either through methods such as textual inversion~\cite{galimage} or fine-tuning ~\cite{ruiz2023dreambooth, kumari2023multi, shi2024instantbooth}.
For segmentation, recent methods establish correspondences between reference and query images to estimate the target concept’s location~\cite{zhang2023personalize, liu2024matcher, tang2024towards, samuel2024wheres}, which is then used to guide prompting of SAM~\cite{kirillov2023segment}. Other solutions have drawn on diffusion models~\cite{tang2024towards, samuel2024wheres}, customized generative models~\cite{sundaram2024personalized}, or cycle-consistency constraints~\cite{cohen2022my}.
In this study, we narrow the focus to personalization in \vlms, aiming to enable recognition, visual question answering, and captioning with references to user-specific concepts.

Early contributions such as MyVLM~\cite{alaluf2024myvlm} and \yollava~\cite{nguyen2024yollava} adapt inversion-based strategies from generative modeling, assigning each object a distinctive latent code, represented either as a concept vector~\cite{alaluf2024myvlm} or a pseudo-token~\cite{nguyen2024yollava, nguyen2025yo}.
To reduce adaptation time, recent efforts incorporate large-scale tuning~\cite{hao2025remember, pi2024personalized, pham2024personalized, an2024mc, das2025training}, improving performance at the cost of extensive pretraining or expensive pairwise-matching.
Most similar to us, \cite{das2025training} leverages the model's implicit knowledge to describe via natural language the reference concept for later matching and~\cite{oh2025repic} employs an RL-based post-training approach for improving \vlm recognition capabilities. However, (i) descriptions generated independently for each concept lack comparative information about visually similar alternatives and (ii) naive application of RL strategies may lead to shortcuts for concept recognition rather than producing a better concept characterization ({see Sec.~\ref{sec:experiments}}).
Unlike prior methods, \ourmethod reframes the problem as a communication game leveraging RL to learn descriptions that uniquely identify the target concept for personalization. 

\paragraph{GRPO-based post-training methods for MLLMs}.
Reinforcement Learning has recently emerged as a powerful paradigm for post-training large language models (LLMs), allowing them to align better with human preferences and downstream objectives.
Approaches such as Reinforcement Learning from Human Feedback (RLHF) and its gradient-based variants have demonstrated remarkable success in improving reasoning and factual consistency in large-scale LLMs \cite{rlhf, summ_rlhf}.
Among these, GRPO (Group Relative Policy Optimization) has become particularly popular due to its stability and efficiency in optimizing non-differentiable objectives during instruction tuning \cite{deepseekmath}.
It enables large models to learn from relative ranking signals among multiple sampled generations, thus avoiding explicit human annotation of reward scores.
Building upon this, methods such as Visual-RFT \cite{liu2025visual} and Vision-R1 \cite{li2025self} have shown that GRPO-guided optimization can significantly improve few-shot performance in vision-centric tasks such as object detection, classification, and visual reasoning.
Most relevant to our work, RePIC \cite{oh2025repic} introduces RL-based post-training for personalizing MLLMs, demonstrating that GRPO can effectively enhance recognition of user-specific visual concepts.
While RePIC focuses on optimizing the reasoning module to predict the correct personalized concept name for a given concept, our approach diverges in its objective.
We employ GRPO to fine-tune the speaker model to generate rich, fine-grained, and state-invariant descriptions.

\paragraph{Learning via communication games.} 
The principles behind multimodal reference games have been studied in various contexts. For instance, \cite{de2017guesswhat} showed how to build an agent that can guess which objects in images an oracle is referring to, akin to referring object segmentation~\cite{kazemzadeh2014referitgame}. Early works on visual dialog~\cite{das2017visual} studied interactions between a model and a human, answering questions given a specific image. Our work takes inspiration from \cite{corona2019modeling}, in which two agents play a guessing game: a speaker communicates an image to a listener using only natural language attributes. Following the theory of mind~\cite{theoryofmind2018rabinowitz,takmaz2023speaking}, the speaker should adapt to what the listener can understand, modifying its language accordingly. Notably, this game has been shown to be applicable for building interpretable classification models~\cite{alaniz2021learning}. In this work, we neither focus on interpretable interactions, nor on full dialogues. Instead, we exploit the abstraction capabilities needed to solve the reference game as a tool for agents to learn how to generate descriptions of concepts tailored for personalization.
\section{\ourmethodlong}
\begin{figure*}[t]
    \centering
    \includegraphics[width=\linewidth]{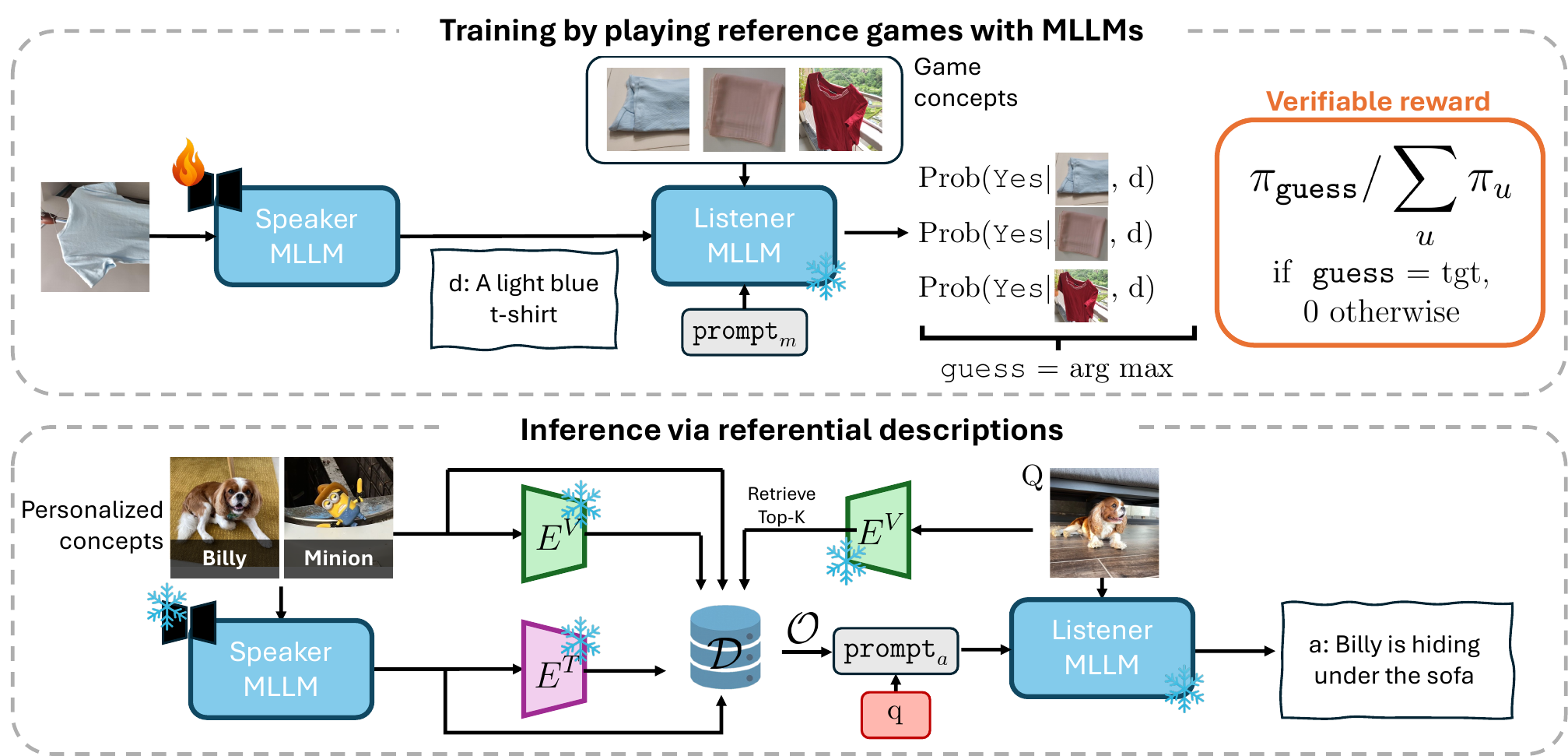}
    \caption{\textbf{\ourmethodlong\ (\ourmethod)}. We revisit personalization through the lenses of a multimodal reference game, where an MLLM should learn how to describe concepts. (Top) During training when acting as a speaker, the MLLM receives as input an image of a concept, producing in output a description. When the MLLM acts as the listener, it receives as input the description from the speaker and a set of images: the goal of the game is to guess to which image the speaker is referring to, assigning the highest matching probability. Based on the outcome of the game, we compute a verifiable reward for RL, updating the speaker. (Bottom) During inference, the speaker outputs descriptions for each of the personalized concepts. These are used as reference for retrieving the personalized concept from a database and aiding MLLM reasoning.}
    \label{fig:method}
\end{figure*}

\label{sec:method}
 We introduce \ourmethodlong, a new method that aims to obtain descriptions tailored for MLLM personalization. We first formalize the personalization problem (\cref{sec:method:formulation}) and present how a reference game can be instantiated on MLLMs for improved personalization using online RL (Sec.~\ref{sec:method:game-mllms}). Finally, we describe how we exploit this newly learned capability during inference (Sec.~\ref{sec:method:inference}). 

\subsection{Problem formulation}
\label{sec:method:formulation}
MLLM personalization aims to inject knowledge of specific personal concepts into a MLLM, in such a way that the model can recognize those personal items and answer queries about them. 
Formally, let us consider the MLLM $\Phi$, as a generative framework, taking as input an image $v\in\mathcal{V}$ and a textual query $t \in \mathcal{T}$, and producing as output a textual response, \ie, $\Phi: \mathcal{V}\times \mathcal{T} \rightarrow \mathcal{T}$. 
In the personalization task, the user provides a set $P=\{p_1, \cdots, p_n\}$ of $n$ \textit{personalized concepts}. Each concept $p\in P$ is a tuple $p=(V_p,t_p)$ composed of a set of reference images $V_p = \{v^1_p, \cdots, v^k_p\}$ and a corresponding name $t_p$, associated to that specific concept. It is important to note that, because these are personal concepts, the MLLM \emph{has not} been previously exposed to any images of $p\in P$ or to the specific names used to reference them. Thus, the set $P$ represents user-specific knowledge that the MLLM must acquire through personalization.
\subsection{Playing reference games with MLLMs}
\label{sec:method:game}

\label{sec:method:game-mllms}
In the personalization scenario of Sec.~\ref{sec:method:formulation}, our goal is to build a MLLM $\Phi_\texttt{speak}$ that produces descriptions useful for personalization.
{{But, how should the speaker behave?}} First, since users may introduce multiple personal concepts over time, the speaker should generate \textit{discriminative} captions \textit{without} requiring access to all personalized concepts simultaneously. 
This ensures that the description of one concept is independent of the others, avoiding the need to update the entire set of descriptions whenever a new concept is added. Second, the speaker should \textit{generalize} to unseen concepts so that it does not require re-training whenever new concepts arrive.
However, supervising the speaker with unique captions poses annotation challenges, as descriptions depend on comparing concepts, a task that is also cognitively challenging for humans.

Multimodal reference games offer a principled framework to meet these requirements. We have two collaborative agents facing an imperfect communication channel: a \textit{speaker} and a \textit{listener}.  Given an image containing a personal concept, the speaker produces a textual description that the listener must use, and based solely on this description, it should identify the concept the speaker is referring to.
Formally, let $C=\{c_1, \cdots, c_n\}$ be the set of \textit{game concepts}. Similarly to a personalized set $P$, to each concept $c\in C$, we associate a name $t_c$ and a set of images $V_c$. Note that, $C$ is a disjoint set of $P$ (\ie, $C \cap P = \emptyset$), as we want the MLLM to generalize its communication capabilities to \textit{arbitrary}, unseen concepts. 
Starting from an image featuring the game concept $c_{tgt}$, the speaker describes the concept it observes, whose description is then used by the listener to identify $c_{tgt}$ from a set of candidate images $V$.
The game is \textit{successful} if the listener guesses $c_{tgt}$, \ie, the image associated to the target concept, and unsuccessful otherwise. Intuitively, successful descriptions re-identify the target concept by capturing features that are both \textit{immutable}, \ie, invariant across views and states, and \textit{distinctive}, meaning that they tell the concept apart from similar ones.

\paragraph{Game setup.} To personalize a given MLLM, we instantiate the game with both $\Phi_\texttt{speak}$ and $\Phi_\texttt{list}$ sharing the same backbone model. The speaker $\Phi_\texttt{speak}$ is further adapted through LoRA~\cite{hulora} parameters $\theta$, which steer it toward producing more discriminative and effective descriptions. 
A game round proceeds as follows: we begin from the game concepts set $C$ and sample a target concept $c_{tgt} \in C$. From its associated image set $V_{tgt}$, one image $v^i_{tgt}$ is provided to the speaker, while a different image $v^j_{tgt}$, with $i \neq j$, is given to the listener to be recognized.
 
\paragraph{Hard positives~\textit{vs.}~Hard negatives.} A key component in our game is how we populate the set of candidates $V$. Since a more challenging game encourages the speaker to be more precise, we construct $V$ using hard examples. On the one hand, hard positives reflect challenging views of the target concept $c_{tgt}$: variations from $v_{tgt}^i$ to $v_{tgt}^j$ may involve changes in illumination, background, or state of the concept (\eg, a blouse being folded or hung). These examples encourage the model to focus on concept-invariant cues while disregarding contextual and appearance variations.
On the other hand, hard negatives acting as distractors need to be discriminated based on their fine-grained, instance-specific attributes, ensuring that the distinction cannot rely on coarse category-level features. To this end, we rely on $U$ concepts from the same semantic class as $c_{tgt}$ and select one image per concept as distractors. For the nature of hard negatives and positives, we identify PerVA dataset~\cite{das2025training} as the most suitable to support our training among other personalization datasets, as it features (i) images of different concepts sharing the same semantic class, ideal for hard negatives, and (ii) diverse views of the same concept with significant appearance changes, ideal for constructing hard positives.
The game round considers recognizing $c_{tgt}$ among candidates $V = \{v^j_{tgt}\} \cup \bar{V}_{tgt}$, with $\bar{V}_{tgt}$ denoting the image set associated to hard negatives.

\paragraph{Game start.}
Given this setup, the speaker produces a description $d=\Phi_\texttt{speak}(v^i_{tgt})$ for the target image, and the listener must determine which image in $V$ the description $d$ refers to. As MLLMs struggle with multi-image inputs~\cite{dingjie2024milebench}, we treat the problem as one with binary output. 

More precisely, for each of the candidates $v_u \in V$ the listener evaluates:
\begin{equation}
\label{eq:matching-probe}
\rho_u = \texttt{Prob}_{\Phi_\texttt{list}}(\texttt{Yes} \mid v_u, d, \texttt{prompt}_m), 
\end{equation}
where $\texttt{prompt}_m$ denotes the prompt used to probe the description-candidate match: \texttt{``Does this description match the provided image? Answer Yes \\or No''}. In other words, \cref{eq:matching-probe} computes the probability that description $d$ refers to the concept $v_{u}$ as evaluated by the probability of the \texttt{Yes} token.

The listener’s final guess is obtained by gathering the information among the candidates, \ie, $\texttt{guess} = \arg\max_u \rho_u$.

\paragraph{Speaker training with contrastive verifiable reward.} We rely on GRPO~\cite{guo2025deepseek}, an online reinforcement learning algorithm to update parameters based on the relative preferences within a set of rollouts, while minimizing the KL divergence between the updated and the reference policy $\pi_\text{ref}$.
The \texttt{guess} outcome of the game provides a verifiable reward signal, defined as:
\begin{equation}
r_\text{match} =
\begin{cases}
\rho_\texttt{guess}/\sum_u \rho_u, & \text{if } \texttt{guess} = tgt, \\
0, & \text{otherwise,} \label{rmatch}
\end{cases}
\end{equation}
where the reward is given by the normalized probabilities of candidates, thus accounting for model uncertainty on its guess. See \suppmat~for more details. 

Hence, the GRPO objective $\mathcal{R}(\theta)$ is:
\begin{equation}
\mathbb{E}
\left[ \frac{1}{S} \sum_{s=1}^S \frac{1}{\lvert d_s\rvert}\sum_{t=1}^{\lvert d_s \rvert} R^s_t(d_s, \theta) - \beta D_{KL}\left(\pi_\theta \mid \mid \pi_\text{ref}\right)\right], 
\end{equation}
where the average is on $S$ rollouts $\{d_s\}_{s=1}^S \sim \Phi_\texttt{speak}(v^i_{tgt})$ when describing an image $v^i_{tgt} \in V_{tgt}$ of a concept $c_{tgt} \in C$. $R^s_t(d_s, \theta)$ is defined as:
\begin{equation}
\text{min}\left(r^s_t(\theta)\hat{A}^s_t, \text{clip}\left(r^s_t(\theta),1 - \epsilon, 1+ \epsilon\right)\hat{A}^s_t\right),  
\end{equation}
where $r_t^s(\theta)$ is the importance sampling ratio and $\hat{A}^s_t$ the normalized advantage accounting for the different rollouts descriptions. 

Success in the reference game translates into the ability to describe the invariant and discriminative features of a concept, aligning the speaker’s behavior with the objectives of concept characterization necessary for MLLM's personalization.
It is important to emphasize that as the training game concepts set $C$ comprises concepts different from the user-defined personal concepts $P$, the model must generalize its learned behavior to unseen user-provided concepts during personalization at inference.

Note that the generalization is over unseen concepts, not unseen \textit{types} of discriminative attributes (\eg, color \vs texture). So the speaker can reuse attribute types already seen in training.

\subsection{Inference via referential descriptions}
\label{sec:method:inference}

Inspired by prior work, we leverage R2P~\cite{das2025training} retrieval-based strategy for inference. Personal concepts $P$ are stored in a database where their visual information is accompanied with referential descriptions from the speaker, and the corresponding concept name. 
Formally, let $v_p \in V_p$ be the visual information of the $p$-th personal concept, with $t_p$ its associated name. We can define $d_p$ as the corresponding speaker-based description $d_p = \Phi_\texttt{speak}(v_p)$. A personal database is constructed as:
\begin{equation}
\mathcal{D} = \{t_p, v_p, d_p, f_p^V, f_p^T\}_{p \in P},
\end{equation}
where $f_p^V = E^V(v_p)$ denotes the visual embedding of the reference image $v_p$ of a CLIP-like~\cite{radford2021learning} encoder $E^V$. For the descriptions we compute textual embeddings $f_p^T = E^T(d_p)$ using the text encoder $E^T$. Starting from the query image $Q$, most relevant concepts according to visual and textual similarity are retrieved:
\begin{equation}
    s_{Qp} = \frac{1}{2}\left(\langle f_p^V, E^V(Q)\rangle + \langle f_p^T, E^V(Q)\rangle\right), \quad p\in P,
    \label{eq:retrieval}
\end{equation}
and the Top-$K$ elements $P^K$ are selected as candidates. Finally, the personalized model reasons on the available concept options $\mathcal{O} = \{(t_p, d_p)\}_{p \in P^K}$ , \ie, possible candidates, to provide a personalized answer. To this end, the referential descriptions of selected candidates are compared to the query image to form a closed-set problem:
\begin{quote}
    $\texttt{prompt}_a(\mathcal{O}, q)$ = ``Consider concept-description pairs $\{\mathcal{O}\}$ and choose among one of the concepts. Please answer $\{q\}$:'',
\end{quote}
where $\texttt{prompt}_a$ is the answer templating function and $q$ refers to the task specific instruction, \ie, recognition, recall or captioning. The final answer evaluates as:
\begin{equation}
a = \Phi_\texttt{list}(Q, \texttt{prompt}_a(\mathcal{O}, q)).
\end{equation}
Hence, the personalized \vlm $\Phi_\texttt{list}$ can produce tailored captions or respond to user-specific queries, conditioned on the user’s prompt. Importantly, if the user explicitly refers to the concept by name $t_p$ in the prompt, the system can directly fetch the associated textual description $d_p$ from $\mathcal{D}$, bypassing the retrieval-and-reasoning stage of the strategy. Please, refer to the \suppmat{} for the implementation details.

\section{Experiments}
\label{sec:experiments}
We evaluate \ourmethod across several established personalization benchmarks, including MyVLM~\cite{alaluf2024myvlm}, \yollava~\cite{nguyen2024yollava}, and PerVA dataset~\cite{das2025training}. We begin by outlining the experimental setting (\cref{sec:experimental-setting}), followed by a comparison against state-of-the-art approaches (\cref{sec:main-comparison}) and motivating experiments on the role of descriptions (\cref{sec:exp:descriptions}). Finally, we present an ablation study analyzing the reward function used in our method and show some qualitative results (\cref{sec:ablation}).

\subsection{Experimental setting}
\label{sec:experimental-setting}

\paragraph{Dataset.} We experiment with three personalization benchmarks: MyVLM~\cite{alaluf2024myvlm}, \yollava~\cite{nguyen2024yollava}, and PerVA~\cite{das2025training}. The MyVLM benchmark defines 29 personalized concepts, each representing user-specific visual concepts such as pets, accessories, or personal belongings, designed to test the model’s ability to recognize and reason on fine-grained visual identity cues. \yollava extends this evaluation to 40 categories covering a broader range of concepts, including people and buildings, thereby offering a more comprehensive benchmark for real-world personalization requests. Finally, we leverage 30 concepts from the PerVA dataset for training the game, while we use the remaining 269 for testing. PerVA stress-tests the personalization on visually ambiguous instances with challenging state changes.

\noindent\textbf{Downstream tasks and metrics.}
We follow prior work~\cite{nguyen2024yollava, hao2025remember,das2025training} and assess the effectiveness of \ourmethod{} across three personalization tasks: \textit{object recognition}, \textit{caption generation}, and \textit{personalized VQA}.
For object recognition, the model must determine whether a query image contains the personalized concept of interest. Positive samples depict the target concept, whereas negative samples contain other concepts from the benchmark. We treat this task as binary classification and report recall (\texttt{Pos.}~$\uparrow$), specificity (\texttt{Neg.}~$\uparrow$), and their balanced average (\texttt{Wtd}~$\uparrow$). For captioning, the objective is to determine whether the model can correctly reference the personalized concept without being explicitly prompted with its name. We report Precision (\texttt{Prec.}~$\uparrow$), Recall (\texttt{Rec.}~$\uparrow$), and their harmonic mean (\texttt{F1}$~\uparrow$), computed per concept and then macro-averaged across all concepts within each dataset.
Finally, for the personalized VQA task, the model answers closed-set questions concerning the personalized concepts and report the answer \texttt{Accuracy} ($\uparrow$). All experiments are averaged on three random seeds.

\noindent\textbf{Baselines.} 
We compare \inlineColorbox{ours}{\ourmethod{}} with several state-of-the-art methods for \vlm personalization. Among training-based approaches, MyVLM~\cite{alaluf2024myvlm} and \yollava~\cite{nguyen2024yollava} leverage \inlineColorbox{ttt}{test-time training} to incorporate new personalized concepts. Instead, RAP~\cite{hao2025remember} (and its Qwen2VL variant, \texttt{RAP-Qwen}) and RePIC leverage large-scale \inlineColorbox{pretrain}{pre-training} and a retrieval-augmented paradigm with an instruction-tuned \vlm to produce personalized responses, with supervised fine-tuning and personalized RL, respectively.
We also compare with R2P~\cite{das2025training}, a \inlineColorbox{tfree}{training-free} approach leveraging textual descriptions for personalization. For fair comparison, we adapted the approach for Qwen2VL removing its costly pairwise matching step, maintaining an equal inference budget across all baselines. Following the VQA task protocol of \cite{das2025training}, as baselines we include \texttt{LLaVA+prompt} and \texttt{GPT-4V+Vprompt}, where the reference description and image are leveraged, respectively.

\subsection{Comparison with state-of-the-art models}
\label{sec:main-comparison}

Table \ref{tab:main_table} reports captioning and recognition performance when comparing \ourmethod to the state of the art.

\begin{wraptable}{r}{0.6\textwidth}
\vspace{-45pt}
\caption{\textbf{Comparison with state of the art}. Recognition and captioning performance ($\uparrow$) on established personalization benchmarks. Captioning metrics are macro-averaged (Precision/Recall/F1).}
\label{tab:main_table}
\scriptsize
\begin{tabularx}{0.6\textwidth}{L{2.1cm}
        *{6}Y}
\toprule
\multirow{2}{*}{\textbf{Method}} 
    & \multicolumn{3}{c}{\textbf{Recognition}} 
    & \multicolumn{3}{c}{\textbf{Captioning}} \\
\cmidrule(lr){2-4} \cmidrule(lr){5-7}
    & Pos. & Neg. & Wtd 
    & Prec. & Rec. & F1 \\ 
\midrule
\multicolumn{7}{c}{\cellcolor{Gray!7}\textbf{MyVLM Dataset~\cite{alaluf2024myvlm}}} \\
\rowcolor{ttt}
MyVLM~\cite{alaluf2024myvlm}                & 96.6 & 90.9 & 93.8  & -    & -    & -    \\
\rowcolor{ttt}
\yollava~\cite{nguyen2024yollava}           & 97.0 & \underline{95.7} & 96.4  & -    & -    & -    \\
\rowcolor{tfree}
\rowcolor{tfree}
R2P-Qwen~\cite{das2025training}             & 86.9 & 95.4 & 91.5  & \underline{94.4}    & \underline{94.1}    & \underline{93.4}    \\
\rowcolor{pretrain}
RAP-LLAVA~\cite{hao2025remember}   & 94.4 & \textbf{98.8} & \underline{96.6}  & 90.1 & 78.7 & 82.0 \\
\rowcolor{pretrain}
RAP-Qwen~\cite{oh2025repic}        & \underline{97.2} & 94.8  & 95.5  & 88.6    & 85.2    & 84.9    \\
\rowcolor{pretrain}
RePIC~\cite{oh2025repic}                    & \textbf{98.1}   & 95.6    & \textbf{96.8}   & 84.2   & 73.2 & 76.7    \\
\rowcolor{ours}
\textbf{\ourmethod (Ours)}                            &  91.5  & 92.0   & 91.8   & \textbf{95.7} & \textbf{95.4}  & \textbf{95.0}    \\
\midrule

\multicolumn{7}{c}{\cellcolor{Gray!7}\textbf{\yollava Dataset~\cite{nguyen2024yollava}}} \\
\rowcolor{ttt}
\yollava~\cite{nguyen2024yollava}           & 94.9 & 89.8 & \underline{92.4}  & -    & -    & -    \\
\rowcolor{tfree}
\rowcolor{tfree}
R2P-Qwen~\cite{das2025training}             & 86.2 & \underline{95.3} & 90.8  & \underline{89.8}    & \underline{87.5}    &  \underline{86.3}   \\
\rowcolor{pretrain}
RAP-LLAVA~\cite{hao2025remember}   & 86.0 & \textbf{99.2} & 92.2  & 83.1 & 70.0 & 76.3 \\
\rowcolor{pretrain}
RAP-Qwen~\cite{oh2025repic}        & \textbf{97.2} & 87.5  & 92.3  &   76.3 & 70.7    & 69.4    \\
\rowcolor{pretrain}
RePIC~\cite{oh2025repic}                    & 91.9   & 91.1   & 91.5   & 79.5    & 57.7    & 62.7   \\

\rowcolor{ours}
\textbf{\ourmethod (Ours)}    & \underline{97.0}   & 92.3   & \textbf{94.7}   & \textbf{91.9} & \textbf{89.3}  & \textbf{88.6}  \\
\midrule

\multicolumn{7}{c}{\cellcolor{Gray!7}\textbf{PerVA Dataset~\cite{das2025training}}} \\
\rowcolor{ttt}
MyVLM~\cite{alaluf2024myvlm}                & 66.0 & 58.5 & 62.2  & -    & -    & -    \\
\rowcolor{ttt}
\yollava~\cite{nguyen2024yollava}           & 75.1 & 69.0 & 72.0  & -    & -    & -    \\
\rowcolor{tfree}
\rowcolor{tfree}
R2P-Qwen ~\cite{das2025training}             & 88.2 & 91.8 & 90.0  & \underline{73.0}    & \underline{67.7}    & \underline{67.3}    \\
\rowcolor{pretrain}
RAP-LLAVA~\cite{hao2025remember}   & 92.9 & 85.2 & 89.0  & 63.8    & 49.5    & 51.8    \\
\rowcolor{pretrain}
RAP-Qwen~\cite{oh2025repic}        & \underline{93.2} & \underline{92.9}  & \textbf{93.1}  &66.0    & 50.4    &52.8    \\
\rowcolor{pretrain}
RePIC~\cite{oh2025repic}                    & 88.8   & \textbf{96.5}   & \underline{92.6}   &62.4    & 44.3    & 48.9    \\
\rowcolor{ours}
\textbf{\ourmethod (Ours)}    & \textbf{94.0}   & 88.3   & 91.1   & \textbf{89.5} & \textbf{87.9}  & \textbf{86.9}    \\
\bottomrule
\end{tabularx}
\vspace{-20pt}
\end{wraptable}

\noindent\textbf{Captioning.} \ourmethod consistently achieves the best results on captioning, on all datasets. On MyVLM, \ourmethod yields substantial gains across all metrics, reaching a 95.0 F1 and markedly surpassing training-based baselines such as RePIC (76.7 \vs 95.0 F1) and RAP-Qwen (+12\% Rec.).
A similar pattern emerges on \yollava, where \ourmethod improves F1 by +16.1\% over the strongest training-based model. Finally, on PerVA, our reference-game 
yields a large margin improvement on F1 over the second-best method (86.9 \vs 67.3 of R2P-Qwen), highlighting the strength of our discriminative and accurate descriptions in characterizing the concepts (see \cref{fig:qualitative}).
Crucially, \ourmethod consistently outperforms R2P-Qwen that leverages textual fingerprints for concept characterization, improving performance across all benchmarks (\eg, +1.6 F1 on MyVLM and +2.3 F1 on \yollava). 
\cref{tab:yollava-categories} offers a breakdown of captioning F1 across different categories of concepts on \yollava (\textbf{L}andmark, \textbf{C}haracter, \textbf{O}bject, \textbf{P}erson and \textbf{A}nimal): while trained on PerVA objects, \ourmethod generalizes to new concepts yielding to best or second-best performance across all categories, including those much different from training ones (\eg, landmark). 
Experiments in the \suppmat{} further demonstrate that \ourmethod is not specific to PerVA. When the same candidate-construction protocol is applied to MC-LLaVA~\cite{an2024mc} concepts, it yields consistent improvements over the zero-shot speaker.

\vspace{2pt}
\noindent\textbf{Recognition.} As for recognition (\cref{tab:main_table}), \ourmethod achieves state-of-the-art Wtd on \yollava (+2.3\% Wtd over the second best) and competitive performance on PerVA, while  we observe a performance gap on MyVLM dataset. 
Our analysis suggests that such limited  performance may stem from hallucinations due to the presence of both the reference image $v_p$ and the query image $Q$ as input. This leads the model to confirm the presence of the described object even when it is absent~\cite{zheng2025modality, yang2025understanding} (see \suppmat{} for a more detailed discussion and a controlled study).
In contrast, captioning requires no reference images at inference, avoiding the visual overload induced by multiple retrieved images and limiting cross-modal interference, with \ourmethod achieving substantial gains on this more challenging task.
Nevertheless, when compared with the description-based R2P-Qwen, our approach consistently achieves better results (\eg, +4.7\% on \yollava, +1.1\% on PerVA). These results confirm the advantage of reference-game-based descriptions over personalization alternatives that rely on textual fingerprints.

\begin{table}[t]
\parbox{.45\linewidth}{
\centering
\caption{\textbf{Comparison with state of the art by category}. F1 Captioning performance ($\uparrow$) on \yollava dataset.}
\label{tab:yollava-categories}
\scriptsize
    \begin{tabular}{lcccccc}
\toprule
      \textbf{Method} & \textbf{L} & \textbf{C} & \textbf{O} & \textbf{P} & \textbf{A} & \textbf{Avg.} \\
\midrule
R2P-Qwen~\cite{das2025training}	& \textbf{100} & 	83.6	& \underline{88.5}	& 75.0	& \textbf{92.1}	& \underline{86.3}\\
RePIC~\cite{oh2025repic}	& 36.4 & 	86.2 & 	75.0 & 	53.7 & 	83.5 & 	62.7\\
RAP-Qwen\cite{hao2025remember}	& 56.3 & 	\textbf{96.7} & 	78.3 & 	\textbf{84.1} & 	70.0	& 76.3\\
\rowcolor{ours}
\textbf{\ourmethod (Ours)} & \underline{98.7} & \underline{93.8} & \textbf{92.2}& \underline{75.7} & \underline{88.2} & 	\textbf{88.6}\\
\bottomrule
\end{tabular}
}
\hfill
\parbox{.45\linewidth}{
\centering
\caption{\textbf{VQA results} in terms of answering accuracy ($\uparrow$) in \yollava~\cite{nguyen2024yollava}.}
    \label{tab:yollava-vqa}
    \begin{tabular}{lc}
\toprule
\textbf{Method} & \textbf{Accuracy} \\
\midrule
\rowcolor{ttt}
Yo'LLaVA~\cite{nguyen2024yollava} & 92.9 \\
\rowcolor{tfree}
GPT-4V + Vprompt~\cite{nguyen2024yollava} & 86.6\\
\rowcolor{tfree}
LLaVA~\cite{nguyen2024yollava}  & 89.9 \\
\rowcolor{tfree}
LLaVA + prompt~\cite{nguyen2024yollava} & 92.5 \\
\rowcolor{tfree}
R2P-Qwen ~\cite{das2025training} & \underline{94.1} \\
\rowcolor{pretrain}
RAP-LLaVA ~\cite{hao2025remember}& {93.2} \\
\rowcolor{ours}
\textbf{\ourmethod (Ours)}  & \textbf{95.3} \\
\bottomrule
\end{tabular}
}
\end{table}

\vspace{2pt}\noindent\textbf{Personalized VQA.}
Table~\ref{tab:yollava-vqa} reports the personalized VQA performance on \yollava.
Our method achieves the highest VQA accuracy of 95.3\%, outperforming the next-best baseline, R2P-Qwen (94.1\%). \ourmethod improves over the large-scale pretrained RAP-LLAVA by 2.1\%.
This improvement highlights the benefit of using accurate and discriminative descriptions generated by our speaker, which provide more unique, informative attributes for the listener MLLM during personalized VQA. 

\subsection{The role of referential descriptions}
\label{sec:exp:descriptions}
\begin{table}[t]
\centering
\caption{\textbf{The effect of descriptions on personalization.} Recognition and captioning performance ($\uparrow$) on MyVLM~\cite{alaluf2024myvlm} and \yollava~\cite{nguyen2024yollava} datasets. Captioning metrics are macro-averaged (Precision/Recall/F1).}
\label{tab:ablation-description}
\scriptsize
\begin{tabularx}{0.8\textwidth}{
L{2cm} 
*{12}Y
}
\toprule
\multirow{3}{*}{\textbf{Method}} 
    & \multicolumn{6}{c}{\textbf{MyVLM}} 
    & \multicolumn{6}{c}{\textbf{Yo'LLaVA}} 
    \\
    \cmidrule(lr){2-7}
    \cmidrule(lr){8-13}
    & \multicolumn{3}{c}{Recognition} & \multicolumn{3}{c}{Captioning} & \multicolumn{3}{c}{Recognition} & \multicolumn{3}{c}{Captioning}\\
    \cmidrule(lr){2-4}\cmidrule(lr){5-7}
    \cmidrule(lr){8-10}\cmidrule(lr){11-13}
    & 
    Pos. & Neg. & Wtd & Prec. & Rec. & F1 & 
    Pos. & Neg. & Wtd & Prec. & Rec. & F1 \\
\midrule
zs-speaker        & 88.4 & 90.2 & 89.3 & 89.9 & 89.4 & 89.3 & 84.7 & \textbf{94.0} & 89.3 & 87.4 & 85.5 & 84.6 \\
listener-training & 89.1 & 90.3 & 89.7 & 92.6 & 92.2 & 92.0 & 88.3 & 92.4 & 90.4 & 89.0 & 87.3 & 86.3 \\
speaker-training  & \underline{91.3} & \textbf{93.0} & \textbf{92.2} & \underline{93.0} & \underline{92.9} & \underline{92.3} & \underline{95.0} & \underline{92.8} & \underline{93.9} & \underline{89.1} & \underline{87.6} & \underline{87.0} \\
\rowcolor{ours}
\textbf{\ourmethod (Ours)} & \textbf{91.5} & \underline{92.0} & \underline{91.8} & \textbf{95.7} & \textbf{95.4} & \textbf{95.0} & \textbf{97.0} & 92.3 & \textbf{94.7} & \textbf{91.9} & \textbf{89.3} & \textbf{88.6} \\
\bottomrule
\end{tabularx}
\end{table}
We conduct experiments to motivate the importance of descriptions and the reference game. \cref{tab:ablation-description} reports personalization results on established MyVLM and \yollava datasets in terms of both recognition and captioning. 

To assess the benefit of enhancing descriptions, we first compare \ourmethod against the pre-trained zero-shot speaker (\texttt{zs-speaker}), which is asked to simply describe the personal concept given the reference image.
The consistent improvements by \ourmethod indicate that refining concept descriptions with RRG enhances personalization in both recognition and captioning tasks (\eg, +5.4 Wtd on \yollava and +5.7 F1 in captioning on MyVLM). 

Furthermore, we investigate whether personalization benefits more from enhancing concept descriptions (training the speaker) or from strengthening concept recognition (training the listener). To this end, we compare \ourmethod against the \texttt{listener-training} baseline in which the personalized model is optimized to recognize objects from zs-speaker descriptions. The results consistently favor prioritizing description quality: \ourmethod yields stronger personalization gains than directly training concept recognition (\eg, +2.1 Wtd in MyVLM, +2.3 in captioning F1 in \yollava). Intuitively, low-quality original descriptions, either lacking discriminative cues or containing irrelevant, distracting attributes, impair listener training.
Notably, the reference game enhances description quality, surpassing a speaker trained on the same data without the game: \ourmethod outperforms \texttt{speaker-training}, where the speaker is optimized with an   MLLM reward model in which the MLLM serves as a verifier, predicting the presence of the target concept conditioned on both the reference image and the generated description (\eg, 95.0 \vs 92.3 F1 on MyVLM and 94.7 \vs 93.9 on \yollava Wtd). We refer the reader to the \suppmat{} for further details on the baselines and for experiments with different listeners: a stronger reward listener improves F1 further, while \ourmethod descriptions continue to outperform zero-shot ones when evaluated with a different inference listener.

Finally, as \ourmethod is a retrieval-augmented strategy, we isolate the impact of improved descriptions on both retrieval and listener reasoning.
To this end, in \cref{fig:retrieval-oracle} we consider a \textit{retrieval oracle} setting in which the target concept is guaranteed to be included among the retrieved candidates. Under this controlled condition, we explicitly evaluate the effect of descriptions on final personalization performance, and compare descriptions produced by a zero-shot speaker (\texttt{zs-speaker}) for the personal concept against those generated by \ourmethod. A consistent gain highlights that enhancing concept descriptions benefits listener reasoning at inference time.

\begin{table}[t]
\begin{minipage}[c]{0.48\linewidth}
	
    \centering
    \vspace{+1.2em}
      \includegraphics[width=\linewidth]{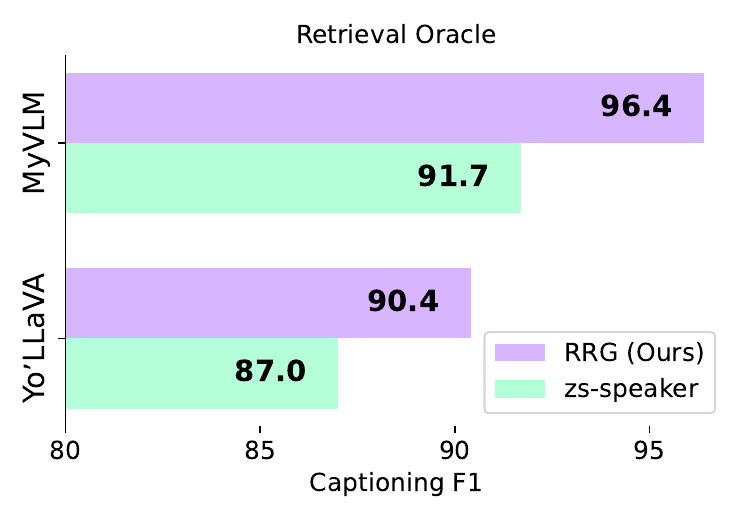}
      \refstepcounter{figure}
      \label{fig:retrieval-oracle}

	\end{minipage}\hfill
	\begin{minipage}[c]{0.48\linewidth}
  \resizebox{\linewidth}{!}{%
  \scriptsize
\begin{tabular}{@{}l cccc@{}}

\toprule
\multirow{2}{*}{\textbf{Method}}
    & \multicolumn{2}{c}{\textbf{MyVLM}}
    & \multicolumn{2}{c}{\textbf{Yo'LLaVA}} \\
    \cmidrule(lr){2-3}
    \cmidrule(lr){4-5}
    & hit@1 & hit@2 & hit@1 & hit@2 \\
\midrule
zs-speaker
    & 85.0 & \underline{96.6} & 84.0 & \underline{97.6} \\
R2P-Qwen~\cite{das2025training}
    & \textbf{92.2} & 96.3 & \textbf{88.1} & 94.3 \\
RePIC~\cite{oh2025repic}
    & \underline{91.5} & 94.9 & 73.5 & 82.5 \\
RAP-Qwen~\cite{hao2025remember}
    & \underline{91.5} & 94.9 & 73.5 & 82.5 \\
\rowcolor{ours}
\textbf{\ourmethod~(Ours)}
    & 88.7 & \textbf{98.5} & \underline{85.4} & \textbf{98.2} \\
\bottomrule
\end{tabular}%
}	
\end{minipage}
\vspace{-1em}
 \captionsetup{labelformat=andfigure}
    \caption{\textbf{The effect of descriptions on reasoning and retrieval.} (Left) Captioning performance ($\uparrow$) on MyVLM~\cite{alaluf2024myvlm} and \yollava~\cite{nguyen2024yollava} under the retrieval oracle setting. (Right) Retrieval performance ($\uparrow$) of different retrieval-based approaches.}
    \label{tab:ablation-retrieval}
\end{table}
Similarly, the retrieval results in \cref{tab:ablation-retrieval} show that more accurate descriptions directly improve retrieval performance, with \ourmethod outperforming the second-best zero-shot speaker in \yollava (98.2 \vs 97.6 hit@2). We also report performance of previous 
retrieval-augmented personalization strategies~\cite{das2025training, hao2025remember, oh2025repic} as reference, with \ourmethod achieving the best hit@2 results on both benchmarks.

\subsection{Additional results}
\label{sec:ablation}

\textbf{Ablation on rewards.} We perform an ablation study to assess the contribution of the proposed reward strategy. Results are reported in \cref{tab:ablation-reward}. The \texttt{binary} baseline, which binarizes the reward based on whether the listener's guess is correct, provides a competitive starting point but underperforms across most metrics. Instead, using the proposed reward $r_\text{match}$  (\cref{rmatch}) yields strong captioning results on both datasets (F1 equal to $95.0$ on MyVLM and to $88.6$ on \yollava), confirming that the normalized probability signal provides a richer training objective than a binary reward. Results are on par in the recognition task where \ourmethod yields +0.8 gain in \yollava but a marginal drop in MyVLM.
\begin{table}[t]
\centering
\caption{\textbf{Ablations on RRG reward.} Recognition and captioning performance ($\uparrow$) on MyVLM~\cite{alaluf2024myvlm} and \yollava~\cite{nguyen2024yollava} datasets. Captioning metrics are macro-averaged (Precision/Recall/F1).}
\label{tab:ablation-reward}
\scriptsize
\begin{tabularx}{0.8\textwidth}{
L{2cm}
*{12}Y
}
\toprule
\multirow{3}{*}{\textbf{Method}} 
     
    & \multicolumn{6}{c}{\textbf{MyVLM}} 
    & \multicolumn{6}{c}{\textbf{Yo'LLaVA}} 
     \\
     \cmidrule(lr){2-7}
     \cmidrule(lr){8-13}
    & \multicolumn{3}{c}{\textbf{Recognition}} & \multicolumn{3}{c}{\textbf{Captioning}} & \multicolumn{3}{c}{\textbf{Recognition}} & \multicolumn{3}{c}{\textbf{Captioning}}\\
    \cmidrule(lr){2-4}\cmidrule(lr){5-7}
    \cmidrule(lr){8-10}\cmidrule(lr){11-13}
    & 
    Pos. & Neg. & Wtd & Prec. & Rec. & F1 & 
    Pos. & Neg. & Wtd & Prec. & Rec. & F1 \\
\midrule
binary & \textbf{92.9} & 91.6 & \textbf{92.3} & 93.5 & 94.2 & 93.4 & 95.4 & \textbf{92.4} & 93.9 & 88.9 & 86.8 & 85.5\\
\rowcolor{ours}
\textbf{$r_\text{match}$ (Ours)} & 91.5 & \textbf{92.0} & 91.8 & \textbf{95.7} & \textbf{95.4} & \textbf{95.0} & \textbf{97.0} & 92.3 & \textbf{94.7} & \textbf{91.9} & \textbf{89.3} & \textbf{88.6} \\
\bottomrule
\end{tabularx}
\end{table}
\begin{figure}
\scriptsize
\begin{tabularx}{\textwidth}{
C{0.17\linewidth}
C{0.25\linewidth}
C{0.25\linewidth}
C{0.25\linewidth}
}
\toprule
& \textbf{zs-speaker} & \textbf{R2P-Qwen} & \textbf{\ourmethod (Ours)} \\
\midrule
\includegraphics[width=0.95\linewidth]{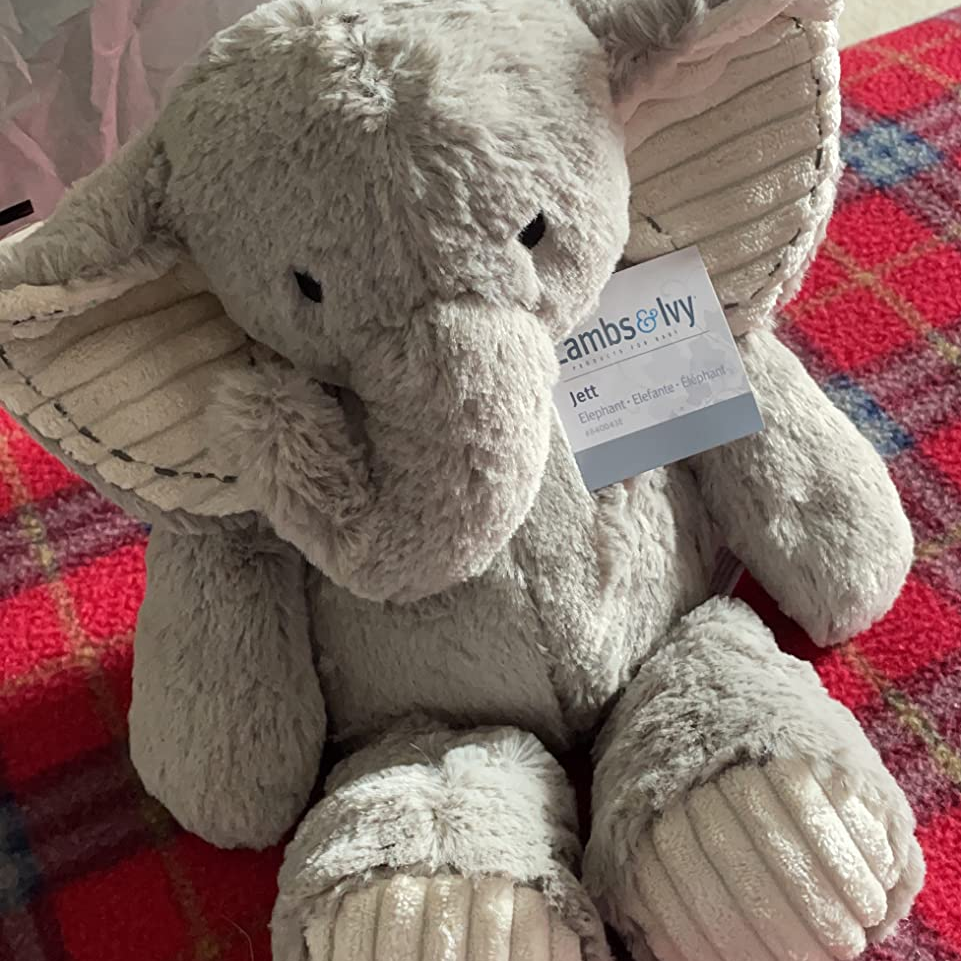} &  {The toy is a gray plush elephant with floppy ears... \hl{is sitting on a red and blue plaid blanket}.} & {Large floppy ears, gray color, black eyes, white tag with 'lambs \& Ivy' branding, \hl{sitting position}.} & {The plush elephant has large, floppy ears, textured feet, and a soft, fuzzy body} \\
\midrule
\includegraphics[width=0.95\linewidth]{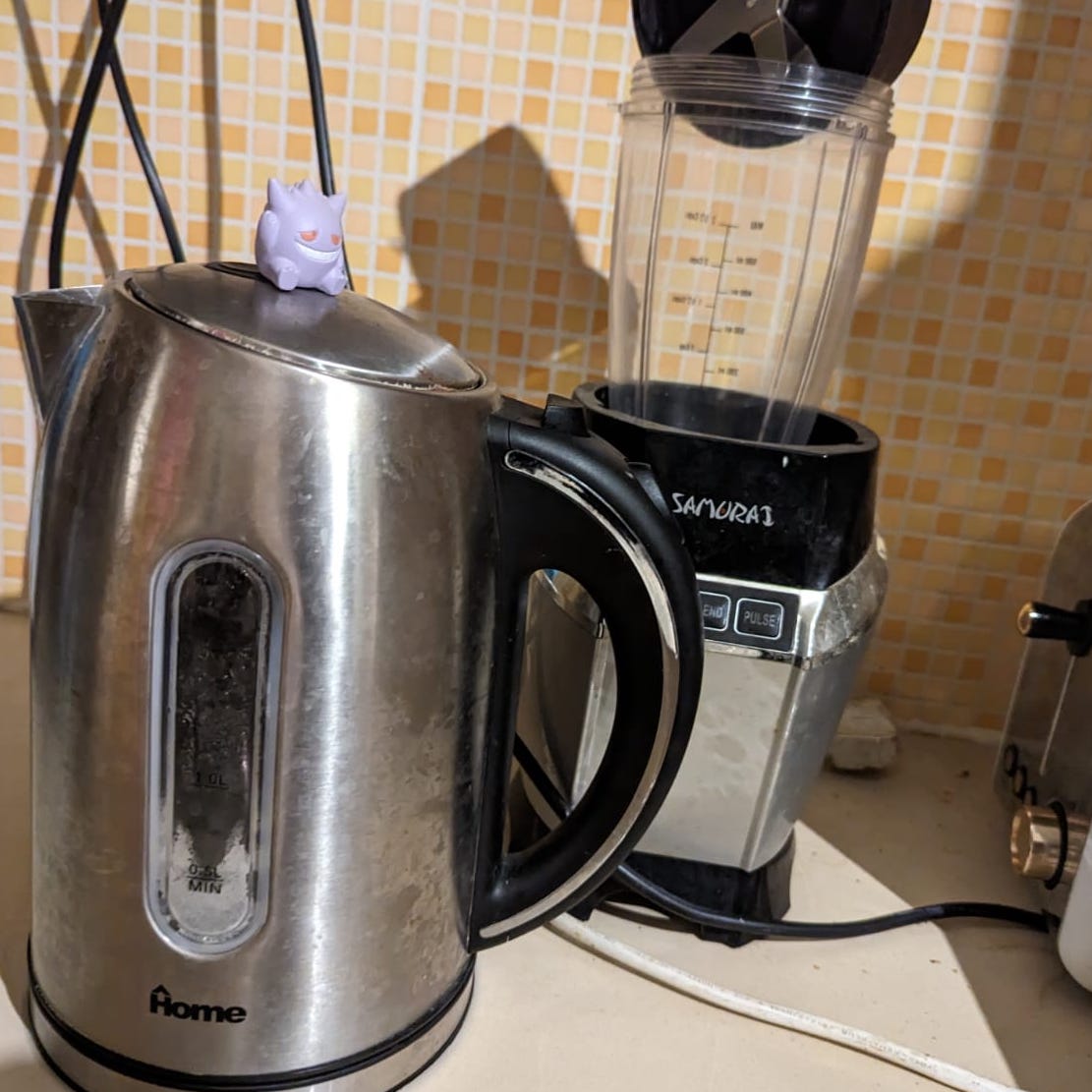} &  {The toy is a small, round, purple creature with a mischievous expression... \hl{positioned on top of a metallic object}.} & {Resembles the character Gengar,
\hl{placed on top of a metal container}.} & {The toy is a small, purple, spiky creature with a mischievous expression and a pink nose} \\
\midrule
\includegraphics[width=0.95\linewidth]{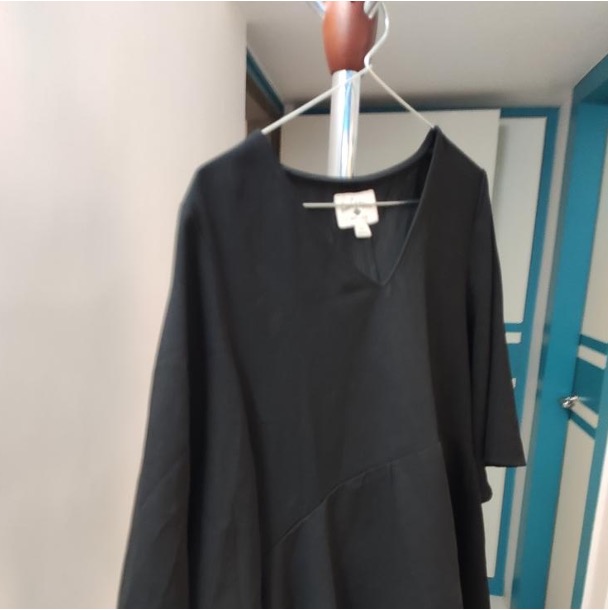} & {The cloth is a black dress with a V-neckline, long sleeve,..,\hl{hanging on a hanger} with a white tag attached.} & {Asymmetrical neckline, full-sleeve design, Smooth texture, \hl{hanging on a hanger}.} & {The black dress features a
distinctive cut with flowing, sheer fabric creating an elegant silhouette.}\\
\bottomrule
\end{tabularx}
\caption{\textbf{Qualitative examples of \ourmethod speaker descriptions.} We compare the descriptions generated by \ourmethod with zs-speaker and R2P-Qwen for three different concepts. Distracting attributes (\eg, background or state-specific) are \hl{highlighted}.}
\label{fig:qualitative}
\end{figure}

\noindent\textbf{Qualitative results on concept descriptions.}
Figure~\ref{fig:qualitative} presents qualitative examples which compare the concept descriptions generated by \ourmethod{} with those produced by the \texttt{zs-speaker} and the training-free \texttt{R2P-Qwen} approach.
As shown, both zs-speaker and R2P-Qwen tend to include unrelated or state-specific features, such as ``hanging on a hanger'' or ``sitting on a plaid blanket'' that vary across images and can distract the listener MLLM during inference.
In contrast, our finetuned speaker consistently avoids such location- or state-dependent artifacts, generating concise descriptions that focus strictly on stable, identity-defining attributes (\eg, ``floppy ears, textured feet'', ``with flowing fabric'') of each personalized concept. Moreover, \ourmethod descriptions add discriminative details that further aid concept recognition, \eg, ``spiky creature'' and ``sheer fabric''.
These invariant descriptions allow the listener to accurately disambiguate among competing personalized concepts, even when the query image differs significantly from the reference ones. More qualitative results are in~\suppmat

\section{Conclusions}\label{sec:conclusions}
We addressed MLLM personalization by proposing a novel learning framework based on a reinforced reference game. In the game, a multimodal large language model learns to communicate a concept to a listener, resulting in personalized descriptions that are increasingly more discriminative and accurate representations of personal concepts.
We validated the advantage of our method, \ourmethod, in comparison to state-of-the-art methods through experiments across three tasks on three benchmark datasets. Our findings show that fine-grained, immutable descriptions significantly enhance downstream recognition, captioning, and VQA performance, without requiring large-scale training or expensive inference-time pairwise reasoning. As future work, we will explore novel joint training strategies for the speaker and listener to enable more efficient and adaptive communication.

\section{Acknowledgement}
We acknowledge ISCRA for awarding this project access to the LEONARDO supercomputer, owned by the EuroHPC Joint Undertaking, hosted by CINECA (Italy). This work was supported by Ministero delle Imprese e del Made in Italy (IPCEI Cloud DM 27 giugno 2022 – IPCEI-CL-0000007), European Union (Next Generation EU), and the EU Horizon ELLIOT (No. 101214398) project. This research has also received funding from the European Union’s Horizon Europe research and innovation actions under grant agreement No 101215032.

%
%
\bibliographystyle{splncs04}
\bibliography{main}
\clearpage
\clearpage
\setcounter{page}{1}
\appendix

\begin{center}
    \Large 
    \textbf{Personalizing MLLMs via Reinforced Multimodal Reference Game} \\
    \vspace{5pt}
    Supplementary Material
\end{center}

In this supplementary material we discuss the implementation details of our strategy (\cref{sec:appx:impl_details}), further detail the considered baselines (\cref{sec:appx:baselines}), and provide additional results (\cref{sec:appx:additional-results}). These include ablations on candidate-set composition, listener robustness and the retrieval score, independence from the PerVA training source, human judgments of description quality, as well as qualitative examples for recognition and captioning tasks. 

\section{Implementation Details}
\label{sec:appx:impl_details}
\paragraph{Base model.}
We use Qwen2-VL-7B~\cite{wang2024qwen2} as the backbone for both the speaker
$\Phi_{\text{speak}}$ and the listener $\Phi_{\text{list}}$.
The speaker is adapted via Low-Rank Adaptation
(LoRA)~\cite{hulora} with rank $r{=}64$ and scaling factor
$\alpha{=}1024$, keeping all other parameters frozen during training.
The listener shares the same base weights but is held frozen throughout.

\paragraph{Training.}
We train the speaker using GRPO~\cite{guo2025deepseek} with a learning
rate of $1{\times}10^{-7}$ and $S{=}4$ rollouts per input image.
We use a per-GPU batch size of~1 with gradient accumulation over 2 steps,
producing an effective batch of 2 concept images per parameter update,
each contributing 4 sampled descriptions.
Training is conducted on the PerVA dataset~\cite{das2025training}, for 1 epoch,
using 30 concepts reserved for the game (disjoint from all evaluation
concepts), with the remaining 269 concepts held out for testing.
All results are averaged over three random seeds.

\paragraph{Reference game setup.}
Each training round presents the speaker with one view $v_{tgt}^{i}$ of
a target concept $c_{tgt}$ and asks it to produce a description that
allows the listener to identify $c_{tgt}$ among $U{+}1$ candidates
$V = \{v^{j}_{tgt}\} \cup \bar{V}_{tgt}$, where $v^{j}_{tgt}$ is a
hard positive (a different view of the \emph{same concept instance},
varying in illumination, viewpoint, or object state) and $\bar{V}_{tgt}$
is a set of $U{=}2$ hard negatives drawn from the same semantic class
as $c_{tgt}$.
Hard negatives are the images \emph{most similar} to $c_{tgt}$
retrieved via a FAISS index on CLIP ViT-L/14-336
({\small\texttt{openai/clip-vit-large-patch14-336}}) features, making
them maximally confusable distractors.
The hard positive is conversely the image of $c_{tgt}$ that is
\emph{least similar} to $v_{tgt}^{i}$ in CLIP space, encouraging the
speaker to produce descriptions that remain discriminative across
appearance variation.

\paragraph{Inference.}
At inference, the speaker generates one referential description per
personalised concept, which is stored in the database $\mathcal{D}$.
Given a query image $Q$, we retrieve the top-$K{=}3$ candidates using the
combined visual-textual similarity score of \cref{eq:retrieval} and pass the
concept-description pairs to the listener. The listener then reasons on the concept description pairs using the chain-of-thought reasoning and selects the most appropriate concept description (containing the name) for the concept image. Once a concept name is assigned, we can use it for various personalized downstream tasks like captioning or VQA.
When the user explicitly refers to a concept by name, the retrieval step
is bypassed, and the stored description is used directly.

\paragraph{Inference-time and storage cost.}
We analyse the practical cost of \ourmethod at inference. Database
construction is a one-time offline step taking ${\sim}5$s per concept.
At query time, retrieval with a FAISS index takes ${\leq}0.1$s per query, while the
final chain-of-thought reasoning step requires ${\sim}3$s. This is
comparable to retrieval-based methods such as RAP~\cite{hao2025remember}
and RePIC~\cite{oh2025repic}, but more efficient than
R2P~\cite{das2025training}, which adds a further ${\sim}8$s per query for
its pairwise verification step that \ourmethod avoids. Storage is
negligible: each concept is represented by a short text description
(${\leq}100$ tokens) alongside its reference embedding, yielding a
lightweight database whose size scales linearly with the number of
personalized concepts.

\subsection{Prompts}
\label{sec:appx:prompt}
\begin{figure}[t]
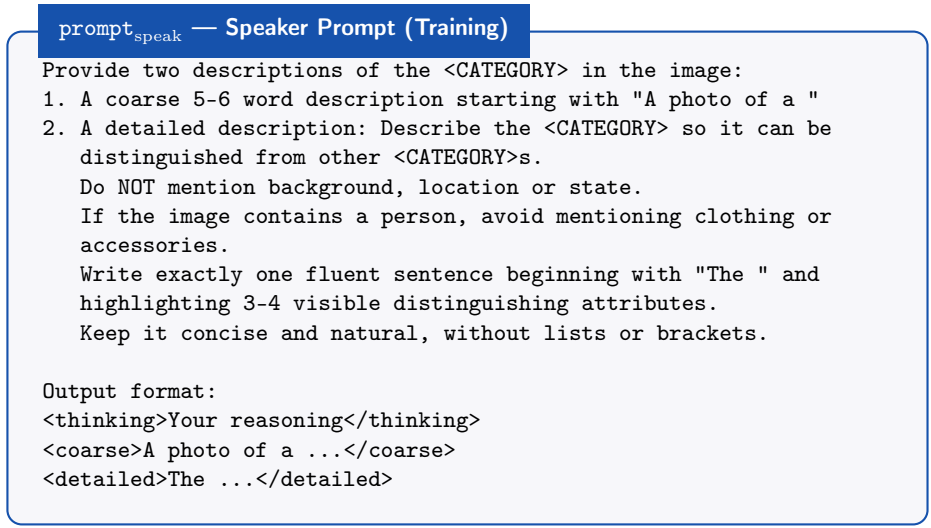

\begin{tcolorbox}[
  speakerstyle,
  title={\(\texttt{prompt}_{\mathrm{speak}}\) \textemdash\ Speaker Prompt (Training)},
]
\begin{lstlisting}[style=promptstyle]
Provide two descriptions of the <CATEGORY> in the image:
1. A coarse 5-6 word description starting with "A photo of a "
2. A detailed description: Describe the <CATEGORY> so it can be
   distinguished from other <CATEGORY>s.
   Do NOT mention background, location or state.
   If the image contains a person, avoid mentioning clothing or
   accessories.
   Write exactly one fluent sentence beginning with "The " and
   highlighting 3-4 visible distinguishing attributes.
   Keep it concise and natural, without lists or brackets.

Output format:
<thinking>Your reasoning</thinking>
<coarse>A photo of a ...</coarse>
<detailed>The ...</detailed>
\end{lstlisting}
\end{tcolorbox}
\caption{%
  \textbf{Speaker prompt} $\texttt{prompt}_{\mathrm{speak}}$.
  Given one image of the target concept, the speaker produces a
  coarse caption and a discriminative detailed description.
  The \texttt{<CATEGORY>} placeholder is filled with the object's
  semantic category (\eg \emph{toy}, \emph{pet animal}) at runtime.
  Structured XML tags allow extraction of both
  description levels during reward computation.
}
\label{fig:prompt_speaker}
\end{figure}

We present the four prompt templates used in our Reinforced Reference
Game framework.
Runtime placeholders are typeset in \texttt{<ANGLE BRACKETS>}.
The speaker (\cref{fig:prompt_speaker}) and listener (\cref{fig:prompt_listener}) prompts are used during GRPO training;
the captioning (\cref{fig:prompt_personalization}) and recognition (\cref{fig:prompt_recognition}) prompts are used at inference only.

\begin{figure}[t]
\begin{tcolorbox}[
  listenerstyle,
  title={\(\texttt{prompt}_{\mathrm{m}}\) \textemdash\ Listener Prompt (Training)},
]
\begin{lstlisting}[style=promptstyle]
Does this description "<REFERENCE DESCRIPTION>" accurately describe the main subject in the image? Answer yes or no.
\end{lstlisting}
\end{tcolorbox}
\caption{%
  \textbf{Listener prompt} $\texttt{prompt}_{\mathrm{m}}$.
  The listener receives the set of images $V$ and the speaker's description for the target concept.
  It answers binary yes/no questions on the presence of the target concept in the images.
  The match reward $r_{\mathrm{match}}$ is computed by comparing
  the prediction against the ground-truth label.
  Placeholder \texttt{<REFERENCE DESCRIPTION>} is filled during run time.
}
\label{fig:prompt_listener}
\end{figure}

\begin{figure}[t]
\begin{tcolorbox}[
  inferstyle,
  title={\(\texttt{prompt}_{\text{cap}}\) \textemdash\ Captioning Prompt (Inference)},
]
\begin{lstlisting}[style=promptstyle]
Reference Descriptions:
- A: <DESCRIPTION OF CONCEPT A>
- B: <DESCRIPTION OF CONCEPT B>
- C: <DESCRIPTION OF CONCEPT C>

Task: Identify which reference matches the subject in Image 1 and generate a personalized caption containing the name of the subject.

Instructions:
1. Compare the subject in Image 1 with each reference description.
2. Focus on distinguishing visual attributes (color, shape,
   pattern, texture).
3. Ignore background, lighting, and pose differences.
4. Select the reference that best matches.
5. Generate a personalized caption containing the name of the subject.

Output format (JSON):
{
  "Reasoning": "Brief comparison of key attributes",
  "Answer": "<one of A, B, C>",
  "Caption": "A personalized caption of the subject in Image 1."
}
\end{lstlisting}
\end{tcolorbox}
\caption{%
  \textbf{Captioning prompt} $\texttt{prompt}_{\text{cap}}$ \textbf{used at inference.}
  The model receives one query image and $K{=}3$ candidate descriptions
  retrieved for the personalised concept.
  It selects the best-matching candidate by reasoning over
  identity-defining visual attributes. Finally, a personalized caption containing the name of the concept is generated.
  This prompt implements the closed-set identification task
  evaluated on the PerVA, MyVLM, and \yollava benchmarks.
}
\label{fig:prompt_personalization}
\end{figure}

\begin{figure}[t]
\begin{tcolorbox}[
  inferstyle,
  title={\(\texttt{prompt}_{\mathrm{rec}}\) \textemdash\ Recognition Prompt (Inference)},
]
\begin{lstlisting}[style=promptstyle]
Reference Description (Image 2):
<REFERENCE DESCRIPTION>

Task: Determine if Image 1 shows the same object as described
for Image 2.

Instructions:
1. Compare the object in Image 1 with the reference description.
2. Focus on identity-defining attributes (not pose, background,
   or lighting).
3. Answer 'Yes' if it is the same object, 'No' if different.

Output format (JSON):
{
  "Reasoning": "Brief comparison of key attributes",
  "Answer": "<Yes or No>"
}
\end{lstlisting}
\end{tcolorbox}
\caption{%
  \textbf{Recognition prompt} $\texttt{prompt}_{\mathrm{rec}}$ \textbf{used at inference.}
  Given a query image and the stored description of the personalised
  concept, the model determines whether the query depicts the exact
  same object instance.
  This binary task is evaluated on MyVLM, \yollava and PerVA as reported
  in the main paper.
  The \texttt{<REFERENCE DESCRIPTION>} placeholder is filled with the description generated by the speaker.
}
\label{fig:prompt_recognition}
\end{figure}
\subsection{Reward Details}
\label{sec:supp_reward}

We provide additional details on our reward scoring mechanism described in \cref{sec:method:game-mllms} of the main paper.

\paragraph{Listener scoring mechanism.}
For each target image $c_{tgt}$, the speaker produces a description $d$ from a reference image of $c_{tgt}$. For each candidate image $v_u \in V$, and the description $d$, the listener answers whether the $v_u$ matches $d$ or not. We calculate the next-token probability that the listener assigns to the \texttt{yes} token ($\rho_u$). The listener's guess is $guess = \arg\max_u\,\rho_u$, \ie the candidate for which it is most confident the description applies. When $guess = tgt$, the reward in \eqref{rmatch} is computed as $\rho_{guess} / \sum_u \rho_u$; otherwise it is zero.

\paragraph{Soft normalized reward vs binary reward.}
A natural choice for the speaker reward is a binary reward, i.e., when $guess = tgt$, the reward is given as $1$ and otherwise $0$. We opt for the soft-normalized reward mainly because it provides a denser reward signal compared to the binary reward. A binary reward simply outputs a score 1 or 0, irrespective of how confident the listener is. Thus an uncertain but correct listener's guess receives the same reward as a confident one. This provides no incentive to the speaker to produce descriptions that are unambiguously discriminative. Conversely, our soft-scoring mechanism takes into account how decisively the listener was able to assign the correct candidate image to the description. If the listener is correct but uncertain our scoring mechanism provides a low score, which motivates the speaker to output a more discriminative and invariant description.

\paragraph{Why is the listener frozen during training?}
We avoid jointly finetuning both the speaker and listener during the reference game mainly for two reasons. The first reason is that finetuning both models would destabilize the training setup. A frozen listener provides the game setup with stable rewards. This would, however, change during joint finetuning as the listener adapts to the speaker descriptions, and thus, there would be no clear way to tell whether the listener model becomes good at guessing due to more discriminative descriptions of the speaker or because the listener adapts better to the speaker's noisy descriptions.
Keeping the listener frozen ensures that $\rho_u$ provides a
\emph{stationary}, semantically grounded reward signal throughout training.

\section{Baselines}
\label{sec:appx:baselines}

We describe the baselines used to study the effect of referential
descriptions in \cref{tab:ablation-description}.

\begin{enumerate}

  \item \texttt{zs-speaker.} The pre-trained speaker model used in a
      zero-shot setting, prompted to describe a personal concept given its
      reference image without any fine-tuning.
      This baseline measures the quality of descriptions produced by a
      capable but non-specialised model before training.
 \item \texttt{R2P-Qwen.}
    R2P~\cite{das2025training} is a training-free personalization approach that
    leverages the MLLM's implicit knowledge to generate natural-language
    \textit{fingerprint} descriptions of personal concepts from their reference
    images. At inference, the stored descriptions are retrieved and matched against the
    query image to identify the concept of interest. During such matching the reasoning model often hallucinates attributes that are not present in the query image. To mitigate this, R2P also uses a pairwise matching approach where it compares the query image and each of retrieved concept's images in an iterative manner. In our comparisons, we adapt R2P by replacing its MiniCPM backbone with Qwen2-VL and removed the pairwise matching step for fair inference budget across all baselines.

 \item \texttt{RAP-LLaVA / RAP-Qwen.}
    RAP~\cite{hao2025remember} is a retrieval-augmented personalization framework
    that pre-trains an MLLM (LLaVA 1.5) on a large-scale dataset of roughly 200k positive–negative concept pairs using supervised fine-tuning.
    At inference, it retrieves the most relevant personal concepts from a database
    using visual similarity and reasons over the retrieved reference images directly. Following~\cite{oh2025repic}, \texttt{RAP-Qwen} replaced the LLaVA backbone with a Qwen 2.5-VL-7B model and was trained on a smaller subset of $2k$ samples instead of the $200k$. All other components remained unchanged.

\item \texttt{RePIC.}
    RePIC~\cite{oh2025repic}, similar to \texttt{RAP-Qwen}, also pretrains on approximately $2k$ subset of positive–negative pairs, but optimizes the model using RL (GRPO) rather than
    supervised fine-tuning.
    It trains the model on multiple objectives based on recognition, grounding as well as captioning. Unlike \ourmethod, which trains the \emph{speaker} to generate more
    discriminative descriptions, RePIC trains the \emph{listener} to recognize
    concepts more reliably from the reference images directly, without improving
    the quality of textual concept descriptions.

\item \texttt{listener-training.} The speaker is held frozen at its
  zero-shot weights while the listener is fine-tuned to adapt to the
  noisy descriptions produced by \texttt{zs-speaker}.
  This isolates whether performance gains can be attributed to a
  better listener rather than better descriptions. The \texttt{listener-training} follows the exact protocol of \ourmethod, with the only difference being that the policy gradients from the rewards are used to update the listener LoRA weights instead of speaker LoRA weights.

  \item \texttt{speaker-training.}
    The speaker is fine-tuned with GRPO while the listener is held frozen,
    but \emph{without} the closed-set reference game.
    Instead of a candidate set $V$, training uses image \emph{pairs}:
    a query image $Q$ containing the target concept $c_{tgt}$ and a
    reference image $v_p$, where $v_p$ is drawn from the same concept as
    $c_{tgt}$ (positive pair) or from a different concept
    (negative pair) with equal probability.
    
    The speaker receives $v_p$ and generates a description $d$.
    The listener is then presented with $(Q,\, v_p,\, d)$
    under the binary prompt $\texttt{prompt}_{list}$ and asked
    whether both images depict the same concept.
    The speaker's reward is binary:
    \begin{equation}
      r =
      \begin{cases}
        1, & \text{if the listener answers correctly,} \\
        0, & \text{otherwise.}
      \end{cases}
    \end{equation}
    This eliminates two components specific to \ourmethod: the use of
 distractors $\bar{V}_{tgt}$ and the soft normalised reward of
    \eqref{rmatch}.
    The baseline therefore isolates whether performance gains from
    \ourmethod arise from game-specific training---discriminative
    distractor sampling and closed-set selection---or from speaker
    fine-tuning alone.
\end{enumerate}

\section{Additional results}
\label{sec:appx:additional-results}
\subsection{Effects of Hard Negatives and Hard Positives}
\begin{table}[t]
\centering
\caption{\textbf{Ablations on hardness of positives and negatives.} Recognition and captioning performance ($\uparrow$) on MyVLM~\cite{alaluf2024myvlm} and \yollava~\cite{nguyen2024yollava} datasets. Captioning metrics are macro-averaged (Precision/Recall/F1).}
\label{tab:ablation-hardness}
\scriptsize
\begin{tabularx}{0.9\textwidth}{
L{2.5cm}
*{12}Y
}
\toprule
\multirow{3}{*}{\textbf{Method}} 
     
    & \multicolumn{6}{c}{\textbf{MyVLM}} 
    & \multicolumn{6}{c}{\textbf{Yo'LLaVA}} 
     \\
     \cmidrule(lr){2-7}
     \cmidrule(lr){8-13}
    & \multicolumn{3}{c}{\textbf{Recognition}} & \multicolumn{3}{c}{\textbf{Captioning}} & \multicolumn{3}{c}{\textbf{Recognition}} & \multicolumn{3}{c}{\textbf{Captioning}}\\
    \cmidrule(lr){2-4}\cmidrule(lr){5-7}
    \cmidrule(lr){8-10}\cmidrule(lr){11-13}
    & 
    Pos. & Neg. & Wtd & Prec. & Rec. & F1 & 
    Pos. & Neg. & Wtd & Prec. & Rec. & F1 \\
\midrule
all-easy & \underline{91}.6 & \underline{92.1} & 91.8 & 94.3 & 93.2 & 92.7 & 96.0 & 93.1 & 94.5 & 90.1 & 88.5 & 88.0 \\
half-easy:half-hard & \textbf{92.5} & \textbf{93.2} & \textbf{92.8} & \underline{94.8} & \underline{94.7} & \underline{93.8} & 96.9 & \textbf{92.7 }& \textbf{94.8} & \underline{91.1} & \underline{89.1} & \underline{88.4} \\
\rowcolor{ours}
\textbf{all-hard (Ours)} & 91.5 & 92.0 & 91.8 & \textbf{95.7} & \textbf{95.4} & \textbf{95.0} &\textbf{97.0} & 92.3 & 94.7 & \textbf{91.9} & \textbf{89.3} & \textbf{88.6} \\
\bottomrule
\end{tabularx}
\end{table}
To study the effect of distractor difficulty on learned description quality,
we ablate the hardness of positives and negatives in the reference game.
\Cref{tab:ablation-hardness} reports recognition and captioning performance
on MyVLM and \yollava.

The \textit{all-easy} variant samples the most-dissimilar concept images from the same
semantic class as easy negatives and the most-similar view of $c_{tgt}$ as
an easy positive,  the inverse of \ourmethod.
The \textit{half-easy:half-hard} variant draws hard positives and negatives 50\% of the
time and easy ones otherwise.

The captioning results follow a clear trend: performance increases
monotonically as the distractors become more challenging, with \ourmethod outperforming
half-easy:half-hard and all-easy on F1 by up to 1.2 and 2.3 points
respectively (MyVLM).
This supports our hypothesis that hard positives and negatives compel the
speaker to discard mutable attributes such as background, pose, and
lighting in favour of permanent identity features, which directly benefits
description-based captioning.

Recognition performance is less sensitive to distractors.
This is expected: recognition is primarily an image-image comparison task
where the listener observes both images directly, making description quality
less decisive (See \cref{sec:supp_qualitative}).
The small recognition gap between conditions (\textless1 point on Wtd in
both datasets) contrasts with the larger and consistent captioning gains,
which is why we adopt all-hard as our final configuration.

\subsection{Candidate-set composition}
\label{sec:appx:candidate-composition}
In addition to the ablation on distractor \emph{hardness}
(\cref{tab:ablation-hardness}), we ablate the \emph{composition} of the
candidate set on MyVLM captioning. \cref{tab:suppl:composition} reports
three settings: using only hard positives (\texttt{h-pos}), only hard
negatives (\texttt{h-neg}), or both (\ourmethod). Using only hard positives
performs worst, as the speaker is never exposed to visually similar
distractors and therefore omits fine-grained discriminative details. Using
only hard negatives does not push the speaker toward view-/state-invariant
properties. Combining both yields the best descriptions, confirming that
hard positives and hard negatives play complementary roles.

\begin{table}[t]
\centering
\scriptsize
\caption{\textbf{Candidate-set composition.} Captioning performance
($\uparrow$) on MyVLM for different candidate-set compositions. Metrics are macro-averaged.}
\label{tab:suppl:composition}
\begin{tabular}{lccc}
\toprule

\textbf{Method} & Prec.  & Rec.  & F1  \\
\midrule
h-pos & 92.7 & 92.7 & 92.1 \\
h-neg & 93.6 & 93.0 & 92.7 \\
\rowcolor{ours}\textbf{\ourmethod (Ours)}    & \textbf{95.7} & \textbf{95.4} & \textbf{95.0} \\
\bottomrule
\end{tabular}
\end{table}

\subsection{Listener reliability and cross-listener transfer}
\label{sec:appx:listener}
Since the listener acts as a frozen reward model during training, two
concerns arise: (i) its own recognition reliability acts as an upper-bound to the quality of
the reward signal, and (ii) the learned descriptions might exploit
single listener's biases. \cref{tab:suppl:listener} addresses both. First,
training with a stronger and newer Qwen2.5-VL-32B listener, whose standalone
reliability is substantially higher (93.7 \vs 89.3 F1 on zero-shot speaker
descriptions), further improves \ourmethod captioning from 95.0 to 95.9 F1
on MyVLM. Second, we test cross-listener transfer by swapping the
\emph{inference} listener for a different model family (InternVL2.5-8B):
\ourmethod descriptions still clearly outperform the zero-shot ones
(83.4 \vs 75.6 F1). Together, these results indicate that the speaker learns
transferable, genuinely discriminative descriptions rather than exploiting a
listener-specific bias.

\begin{table}[t]
\centering
\scriptsize
\caption{\textbf{Listener reliability and cross-listener transfer.}
Captioning F1 ($\uparrow$) on MyVLM.}
\label{tab:suppl:listener}
\begin{tabular}{l C{3cm} c}
\toprule
\textbf{Method (training listener)} & Inference listener & F1 \\
\midrule
zs-speaker          & Qwen2-VL-7B    & 89.3 \\
zs-speaker         & Qwen2.5-VL-32B & 93.7 \\
\rowcolor{ours}\textbf{\ourmethod (Qwen2-VL-7B)}     & Qwen2-VL-7B    & 95.0 \\
\rowcolor{ours}\textbf{\ourmethod (Qwen2.5-VL-32B)}  & Qwen2-VL-7B    & \textbf{95.9} \\
\midrule
zs-speaker          & InternVL2.5-8B & 75.6 \\
\rowcolor{ours}\textbf{\ourmethod (Qwen2-VL-7B)}     & InternVL2.5-8B & \textbf{83.4} \\
\bottomrule
\end{tabular}
\end{table}

\subsection{Independence from the PerVA training source}
\label{sec:appx:perva-dependency}
\ourmethod requires a training source providing (i) multiple views/states
of the same concept (for hard positives) and (ii) several instances of the
same semantic class (for hard negatives), but it is not tied to PerVA
specifically. To show this, we apply the identical candidate-construction
protocol to MC-LLaVA~\cite{an2024mc}
character and person concepts, and evaluate on the MyVLM benchmark (Object
and Animal concepts). As reported in \cref{tab:suppl:mcllava}, \ourmethod
improves over the zero-shot speaker on all captioning metrics, confirming
that the approach generalizes beyond PerVA. The somewhat larger gains
obtained when training on PerVA are expected, as PerVA offers particularly
challenging multi-view/state variations and same-category hard negatives
that yield a stronger training signal.

\begin{table}[t]
\centering
\scriptsize
\caption{\textbf{Independence from the PerVA training source.} Captioning
performance ($\uparrow$) on MyVLM when training \ourmethod on
MC-LLaVA~\cite{an2024mc} concepts instead of PerVA. Metrics are macro-averaged.}
\label{tab:suppl:mcllava}
\begin{tabular}{lccc}
\toprule
\textbf{Method} & Prec.  & Rec. & F1  \\
\midrule
zs-speaker & 89.9 & 89.4 & 89.3 \\
\rowcolor{ours}\textbf{\ourmethod  (Ours)}        & \textbf{94.8} & \textbf{94.4} & \textbf{94.1} \\
\bottomrule
\end{tabular}
\end{table}

\subsection{Human study on description quality}
\label{sec:appx:human-study}
All metrics in the main paper assess descriptions \emph{indirectly}, through
downstream recognition, captioning, and VQA. To assess description quality
\emph{directly}, we conducted a human study with 12 participants, collecting
281 annotations. Participants rated descriptions on a 1--5 Likert scale along
three dimensions: \textbf{D}iscriminativeness (does the description single
out the concept among similar ones?), \textbf{C}ontext-invariance (does it
avoid background/state/context?), and attribute \textbf{S}tability (are the
mentioned attributes invariant across views?).
As shown in \cref{tab:suppl:human}, \ourmethod is preferred on all three
axes, confirming that its gains are not merely reflected in downstream tasks
but also in human-perceived description quality.

\begin{table}[t]
\centering
\scriptsize
\caption{\textbf{Human study on description quality.} Mean Likert ratings
($\uparrow$, 1--5) over 281 annotations from 12 participants.}
\label{tab:suppl:human}
\begin{tabular}{lccc}
\toprule
\textbf{Method} & Discriminativeness  & Context-invariance  & Stability  \\
\midrule
zs-speaker & 3.70 & 3.48 & 3.70 \\
\rowcolor{ours}\textbf{\ourmethod (Ours)}          & \textbf{4.05} & \textbf{4.27} & \textbf{4.31} \\
\bottomrule
\end{tabular}
\end{table}

\subsection{Ablation on the retrieval score}
\label{sec:appx:retrieval-ablation}
We ablate the two components of the retrieval score in \cref{eq:retrieval}
(textual-only and the full multimodal score) on MyVLM. On the textual
branch, \ourmethod descriptions improve retrieval over the zero-shot speaker
(Hit@1/2: $76.4/91.8 \rightarrow 79.0/92.8$), showing that description
quality directly affects retrieval. The full multimodal score performs best
by combining global visual similarity with discriminative textual cues, with
multimodal \ourmethod reaching $88.7/98.5$ Hit@1/2 ($+5.7/+3.3$ over the
multimodal zero-shot speaker). This motivates the multimodal design of
\cref{eq:retrieval}.

\begin{table}[t]
\centering
\scriptsize
\caption{\textbf{Ablation on the retrieval score of \cref{eq:retrieval}.}
Retrieval Hit@1/Hit@2 ($\uparrow$) on MyVLM for the textual-only branch and
the full multimodal score.}
\label{tab:suppl:retrieval-eq6}
\begin{tabular}{lC{2.4cm}cc}
\toprule
\textbf{Retrieval score} & Descriptions & Hit@1  & Hit@2  \\
\midrule
Text-only  & zs-speaker & 76.4 & 91.8 \\
Text-only  & \ourmethod          & 79.0 & 92.8 \\
Multimodal & zs-speaker & 83.0 & 95.2 \\
\rowcolor{ours} Multimodal & \ourmethod          & \textbf{88.7} & \textbf{98.5} \\
\bottomrule
\end{tabular}
\end{table}

\subsection{Qualitative Analysis}
\label{sec:supp_qualitative}
We present two sets of qualitative examples from our method. The first qualitative (\cref{fig:supp_recognition_errors}) illustrates the failure modes in our recognition task, while the second qualitative (\cref{fig:supp_description_comparison}) illustrates the effect of description quality on listener accuracy.
\subsection*{Hallucination in Recognition}
\begin{figure}[t]
  \centering
  \includegraphics[width=\linewidth]{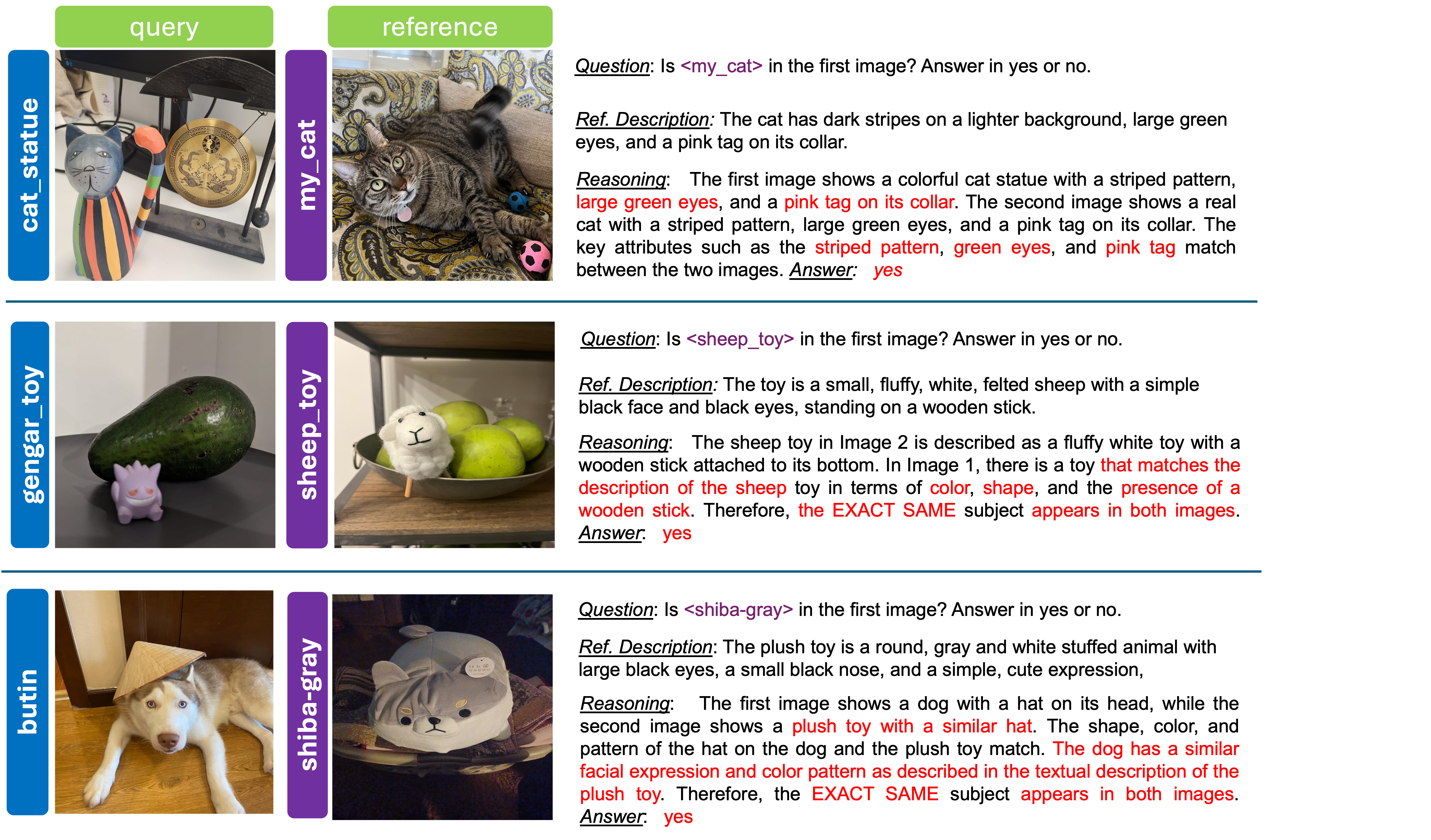}
  \caption{%
    \textbf{Recognition failures driven by text-anchored hallucination.}
    Each row shows a query image (blue label) paired with a reference
    concept image (purple label); all three are \emph{negative} pairs
    (different objects), yet the model answers ``yes'' in every case.
    Highlighted phrases in the reasoning reveal how the model projects
    attributes from the reference description onto the query image
    (\eg, attributing ``striped pattern, green eyes, pink tag'' to a
    painted cat statue; ``presence of a wooden stick'' to a Gengar
    figurine) to manufacture spurious evidence of a match.
  }
  \label{fig:supp_recognition_errors}
\end{figure}
\Cref{fig:supp_recognition_errors} shows recognition failures drawn
from the MyVLM and \yollava benchmarks.
In each case the model is given a query image (blue label), a
reference concept image (purple label) and a description belonging to the reference image. Given these inputs the model must answer whether both depict the same object.
Despite the two images showing \emph{different} objects, the model
consistently answers ``yes''.

Examining the chain-of-thought reasoning revealed that apart from the purely visual confusion while comparing two images on fine-grained attributes (row 3, model hallucinates that shiba-gray is also wearing a hat), the model also
relies on the textual description associated with the reference
concept and then re-reads the query image \emph{through the lens of
that description}.
This leads it to hallucinate matching attributes that are simply not
present in the query.
In the first row, a painted \texttt{cat\_statue} is claimed to exhibit
``large green eyes and a pink tag on its collar'' because the
reference \texttt{my\_cat} possesses those features while in the second row, a purple \texttt{gengar\_toy} is said to have ``the
presence of a wooden stick'' to match the \texttt{sheep\_toy} description.

This behaviour is particularly pronounced on MyVLM, where the concept
descriptions are richly detailed, providing more anchoring text for
the hallucination mechanism to exploit.
The qualitative failures in \cref{fig:supp_recognition_errors} are
corroborated quantitatively in \cref{tab:suppl:hallucination}, which varies
the input provided to the listener at inference on the MyVLM recognition
task. With a text-only input ($d_p$, the stored description), the listener
attains the highest positive recall but the lowest negative accuracy: a
pronounced ``yes'' bias in which rich descriptions are projected onto
unrelated query images, producing false positives. This is consistent with
language-dominance findings in MLLMs~\cite{zheng2025modality, yang2025understanding, zhou2023analyzing}.
Multimodal inputs (\texttt{zs-speaker} and \ourmethod, which additionally see
the reference image) mitigate this bias, but do not fully remove it:
differences between the reference and the query can instead be misread as
evidence of concept absence, lowering positive accuracy. This explains the
recognition gap on MyVLM discussed in the main paper, while
captioning, which uses no reference image at inference, is unaffected.

\begin{table}[t]
\centering
\scriptsize
\caption{\textbf{Recognition hallucination on MyVLM.} Effect of the listener
input on recognition ($\uparrow$).}
\label{tab:suppl:hallucination}
\begin{tabular}{lccc}
\toprule
\textbf{Method} & Pos. $\uparrow$ & Neg. $\uparrow$ & Wtd. $\uparrow$ \\
\midrule
$d_p$ (zs-speaker, text-only) & \textbf{95.7} & 89.0 & 92.4 \\
zs-speaker (multimodal)       & 88.4 & 90.2 & 89.3 \\
\rowcolor{ours}\textbf{\ourmethod (multimodal)}                & 91.5 & \textbf{92.0} & 91.8 \\
\bottomrule
\end{tabular}
\end{table}
\subsection*{Effect of Description Quality on Listener Accuracy}
\begin{figure}[t]
  \centering
  \includegraphics[width=\linewidth]{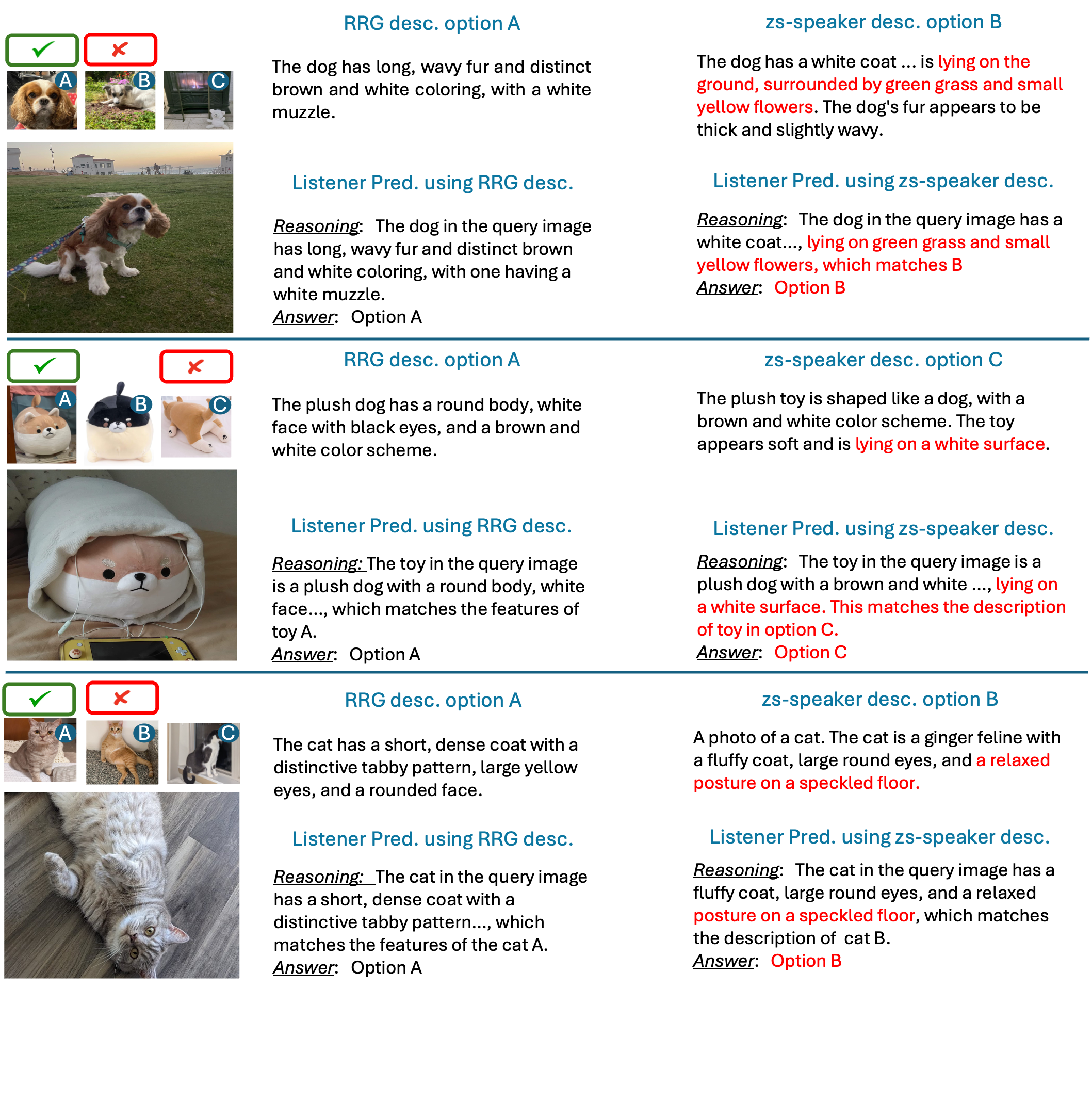}
  \caption{%
    \textbf{\ourmethod descriptions enable correct listener predictions;
    zero-shot speaker descriptions do not.}
    Each row presents a query image and three candidate concepts
    (A, B, C), where A is always the ground truth.
    \emph{Middle column}: RRG description for option~A and
    the resulting listener prediction (correct in all cases).
    \emph{Right column}: zero-shot speaker (zs-speaker) description
    for the incorrectly chosen option, highlighted in red, and the
    resulting listener prediction (wrong in all cases).
    The zs-speaker descriptions contain transient state and location
    details (\eg, pose, background surface, surrounding objects) that
    incidentally match the query image, misleading the listener into
    selecting the wrong concept.
    RRG training steers the speaker away from such details, producing
    descriptions grounded in permanent, identity-defining attributes
    that generalise across viewpoints and contexts.
  }
  \label{fig:supp_description_comparison}
\end{figure}
\Cref{fig:supp_description_comparison} illustrates how description
quality influences listener accuracy.
Each row shows a query image alongside three retrieved candidate
options (A, B, C).
The green tick marks the ground-truth concept while the red cross marks the option incorrectly selected by the listener when given zero-shot speaker (zs-speaker) descriptions.

The critical difference is highlighted in red: zs-speaker
descriptions tend to include \emph{transient} state and location
details, ``sitting on a blue dog bed with a red and white polka-dot
interior'', ``lying on a white surface'', ``a relaxed posture on a
speckled floor'', rather than \emph{invariant} attributes.

Since these transient features are also visible in the query image
(the dog in the query \emph{is} lying on a surface; the cat
\emph{is} on a floor), the listener latches onto the spurious
background match and selects the wrong concept.
RRG descriptions, shaped by the reference-game reward, avoid such
details in favour of permanent visual attributes (\eg, coat colour,
facial markings, body shape), enabling the listener to reliably
distinguish the correct concept from distractors.
\end{document}